\documentclass[letterpaper, 10 pt, conference]{ieeeconf}  % Comment this line out if you need a4paper

\IEEEoverridecommandlockouts                              % This command is only needed if
\makeatletter
\let\NAT@parse\undefined
\makeatother
\usepackage[numbers,sort&compress]{natbib}

\usepackage{graphics} % for pdf, bitmapped graphics files
\usepackage{mathptmx} % assumes new font selection scheme installed
\usepackage{times} % assumes new font selection scheme installed
\usepackage{amsmath} % assumes amsmath package installed
\usepackage{amssymb}  % assumes amsmath package installed

\usepackage{subfigure}
\usepackage{graphicx}
\usepackage{booktabs}
\usepackage{threeparttable}
\usepackage{multirow}

\usepackage{hyperref}
\hypersetup{
  colorlinks=true, %Colours links instead of ugly boxes
  urlcolor=blue,   %Colour for external hyperlinks
  linkcolor=black, %Colour of internal links
  citecolor=black  %Colour of citations
}

\usepackage{balance}

\usepackage{rpm_SIunits}
\usepackage{rpm_acronyms}
\usepackage{rpm_math}
\usepackage{rpm_misc}

\usepackage{soul,color}

\usepackage[font=small,labelfont=bf]{caption}

\usepackage{xcolor}

\usepackage{lipsum}

\title{\LARGE \bf Asynchronous Multiple LiDAR-Inertial Odometry\\using Point-wise Inter-LiDAR Uncertainty Propagation
}
\author{Minwoo Jung$^{1}$, Sangwoo Jung$^{1}$ and Ayoung Kim$^{1*}$
\thanks{$^{1}$ M. Jung, S. Jung and A. Kim are with the Dept. of Mechanical Engineering, SNU, Seoul, S. Korea {\tt\footnotesize [moonshot, dan0130, ayoungk]@snu.ac.kr}}%
\thanks{$^\dagger$This research was funded by the Korea MOLIT (23SMIP-A158708-04).}%
}

\begin{document}

\maketitle
\thispagestyle{empty}
\pagestyle{empty}

\begin{abstract}
In recent years, multiple \ac{LiDAR} systems have grown in popularity due to their enhanced accuracy and stability from the increased \ac{FOV}. However, integrating multiple LiDARs can be challenging, attributable to temporal and spatial discrepancies. Common practice is to transform points among sensors while requiring strict time synchronization or approximating transformation among sensor frames. Unlike existing methods, we elaborate the inter-sensor transformation using \ac{CT} \ac{IMU} modeling and derive associated ambiguity as a point-wise uncertainty. This uncertainty, modeled by combining the state covariance with the acquisition time and point range, allows us to alleviate the strict time synchronization and to overcome \ac{FOV} difference. The proposed method has been validated on both public and our datasets and is compatible with various LiDAR manufacturers and scanning patterns. 
We open-source the code for public access at \url{https://github.com/minwoo0611/MA-LIO}.
\end{abstract}

%This paper tackles these challenges by incorporating B-spline interpolation for asynchronous sensor streams, alleviating the strict time synchronization. To this end, point-wise uncertainty is modeled by combining the state covariance with the acquisition time and point range.

\section{INTRODUCTION \& RELATED WORKS}
\label{sec:intro}

Over the last decades, robot navigation using \ac{LiDAR} has made substantial advances in localization and map construction. Although existing methods mostly solve for a single \ac{LiDAR} system, limited \ac{FOV} and occlusion lead to a need for multiple \ac{LiDAR}s. When integrating multiple \ac{LiDAR}s in a complementary configuration, two major challenges impede naive integration, namely temporal and spatial discrepancy.

%FIGURE
\begin{figure}[!t]
     \centering
     \subfigure[Multi-LiDAR datasets]{
         \centering
         \includegraphics[width=0.8\linewidth]{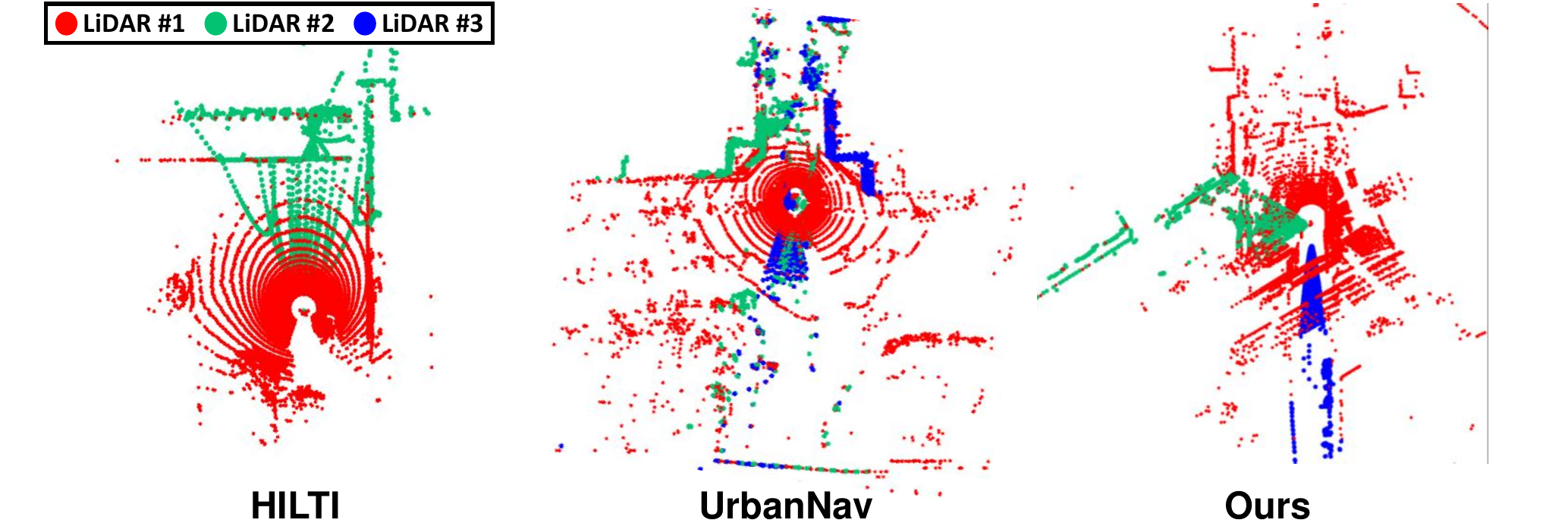}}
         \vspace{-2mm}
    \subfigure[Constructed map using multi-LiDAR]{
         \centering
         \includegraphics[width=0.8\linewidth]{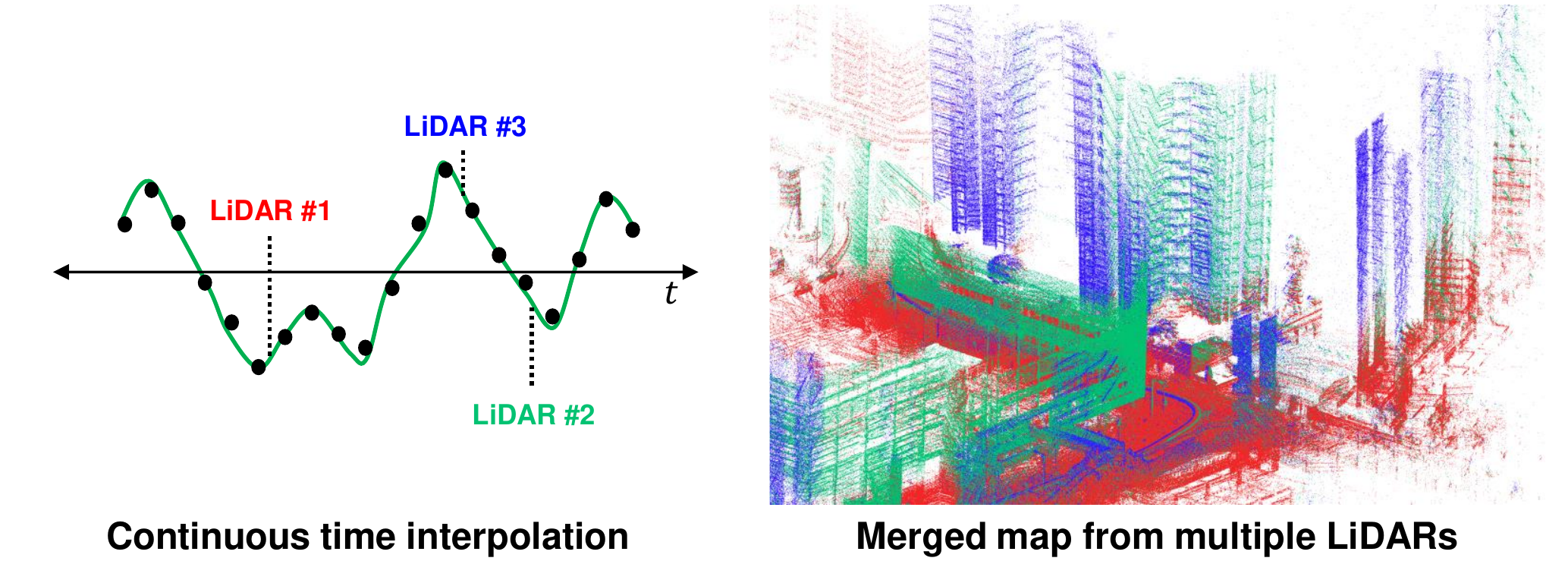}
         }

    \subfigure[Two types of uncertainties]{
         \centering
         \includegraphics[width=0.8\linewidth]{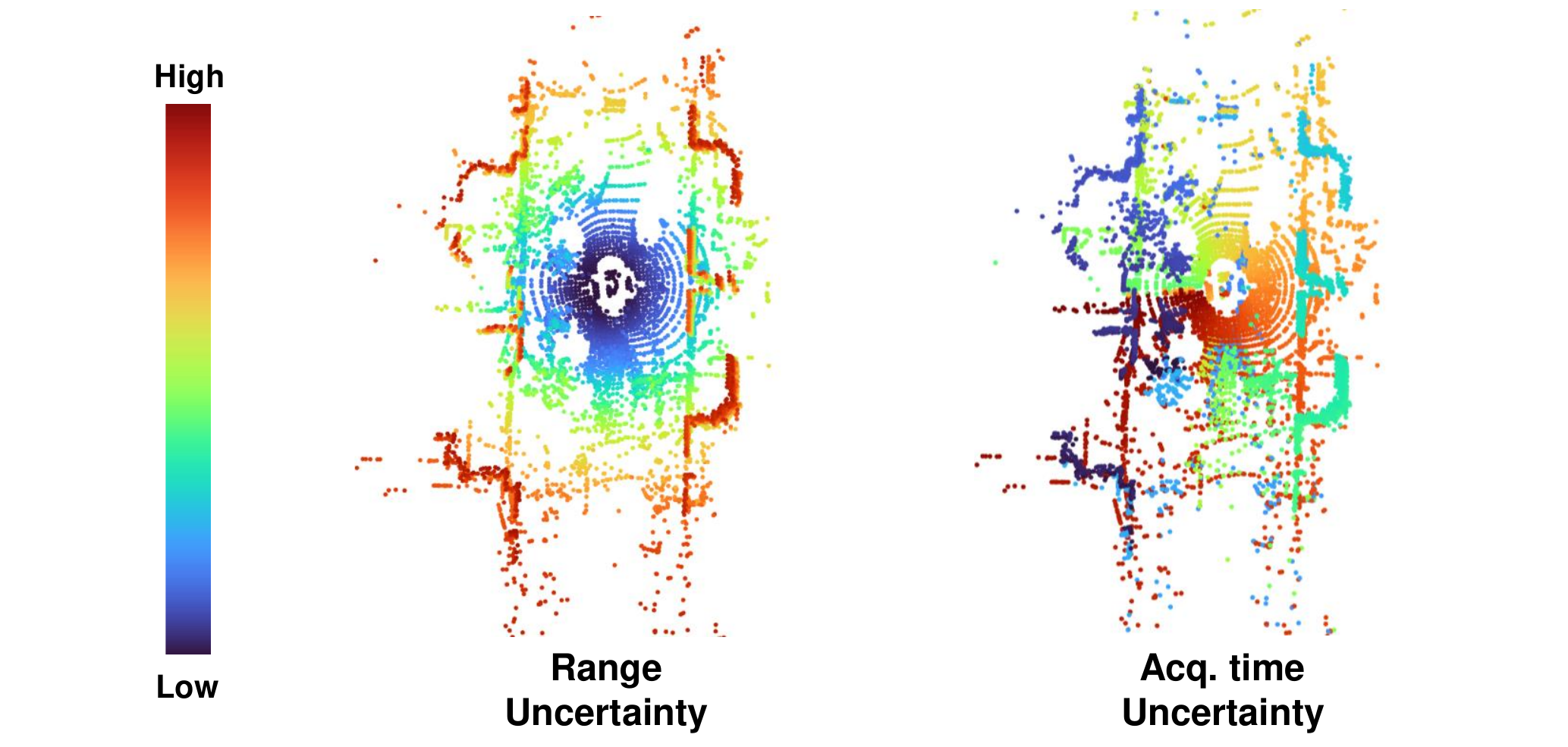}}
        \caption{(a) Example of FOV difference in multi-LiDAR datasets (b) \ac{CT} IMU interpolation enable us to merge points accurately, thus substantially increasing net FOV in the accumulated map. (c) Two types of uncertainties considered in this paper.}
        \label{fig:main}
    \vspace{-6mm}
\end{figure}

%FIGURE

\textbf{(\textit{i}) Synchronization: } It is straightforward and enticing to secure strict time synchronization among sensors to obtain all measurements at the same time for easy integration. M-LOAM \cite{jiao2021robust} proposed the multiple LiDAR odometry for carefully synchronized LiDAR. In practice, sensors can be synchronized by \ac{PPS} via external hardware for transmitting the \ac{PPS} signal. Another option is \ac{PTP}, yet not all manufacturers support it, thus requiring sensor combinations made by the same manufacturers.

Due to this sophisticated setting for synchronization, some asynchronous public datasets \cite{helmberger2022hilti, hsu2021urbannav, nguyen2021ntu, reinke2022locus} require software solutions to handle temporal discrepancies. For example, LOCUS \cite{palieri2020locus} assumed synchronization to merge pointclouds. Later in LOCUS 2.0 \cite{reinke2022locus}, they discarded delayed scans for improved robustness accepting induced information loss. \citeauthor{lin2020decentralized} \cite{lin2020decentralized} applied decentralized Extended Kalman Filter for asynchronous Livox LiDARs; however, their small \ac{FOV} could induce the error in degenerate environments when processing each LiDAR separately. \citeauthor{nguyen2021miliom} \cite{nguyen2021miliom} and \citeauthor{wang2022simultaneous} \cite{wang2022simultaneous} exploited the IMU to compensate for temporal discrepancies. The idea of utilizing IMU was affordable; still, the error originating from discrete propagation remains. Some researchers suggested using optimization in continuous-time formulation \cite{lv2021clins, mueggler2018continuous}. While this approach can estimate the entire trajectory at any point in time, intensive computational burden hindered real-time feasibility.

%This is particularly true for multi-LiDAR systems, which typically involve more measurements and computational resources.

%While these methods may be convenient for addressing synchronization issues, they yield information loss from discarded scans and cannot compensate for the small temporal discrepancy.

In contrast, our method utilizes all available information, including asynchronous scans, and avoiding linear approximation. We decided not to include the continuous term in the optimization phase for a real-time process. Instead, B-Spline interpolation is leveraged to estimate the trajectory for each LiDAR measurement, enabling a computationally efficient solution for multi-LiDAR SLAM in real-world environments.

\textbf{(\textit{ii}) Spatial discrepancy: } Another solution for this temporal discrepancy is to apply scan matching and compute necessary correction. However, spatial discrepancy induced from different scanning patterns and \ac{FOV} among LiDARs hinders scan matching in overcoming the temporal discrepancy. Specifically, sparsity and \ac{FOV} variance from non-repetitive scanning patterns \cite{helmberger2022hilti} and obliquely installed spinning pattern \cite{hsu2021urbannav}, may yield a little overlapping area among sensors.

\textbf{(\textit{iii}) Uncertainty propagation: } From both temporal and spatial discrepancy, we inevitably accumulate ambiguity during the projection of the points among sensors. Regarding this issue, proper uncertainty modeling could capture the transferred ambiguity among multiple \ac{LiDAR}s. The usage of uncertainty in \ac{LiDAR} has been extensively studied in recent years, mostly focusing on learning-based methods \cite{Uncertainty1, Uncertainty2, Uncertainty3, Uncertainty4}.

Yet, like ours, some model-based approaches have been presented. For example, \citeauthor{wang2022simultaneous} \cite{wang2022simultaneous} calculated the weight for LiDAR residual based on the difference between IMU and LiDAR odometry. Whereas all points were treated equally in \cite{wang2022simultaneous}, we assign point-wise uncertainty by considering their range and acquisition time. This is particularly important to handle both ambiguities induced by the temporal and spatial discrepancy. M-LOAM \cite{jiao2021robust} propagated the uncertainty in each point by using the extrinsic covariance and the state covariance. This approach is similar to ours in that utilizing state covariance and point-wise uncertainty propagation \cite{barfoot2014associating}. Differing from M-LOAM, our method requires no inter-LiDAR overlap for extrinsic covariance update and thus is more generic in handling point-level uncertainty. In \cite{9981157}, authors considered the variance of normal direction using uncertainty to deal with the uneven terrain but without incorporating acquisition time and range ambiguity.

\begin{figure}[!t]
	\centering
		\includegraphics[width=1\columnwidth]{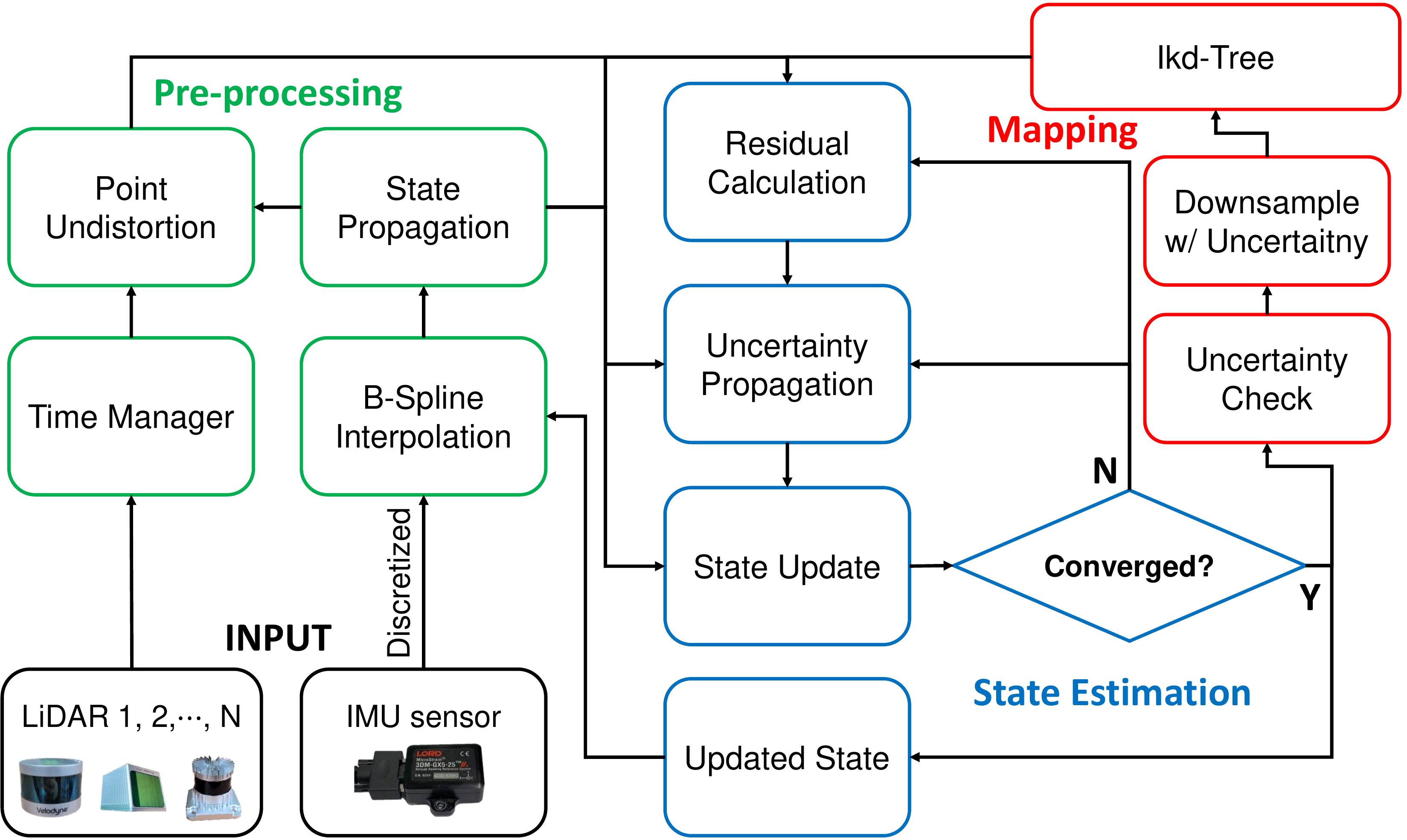}
  	\caption{The proposed method consists of three modules: preprocessing, state estimation, and mapping. In the preprocessing stage, the points from each LiDAR are corrected for distortion and merged by B-spline interpolation. The state estimation stage involves point-wise uncertainty propagation and the application of an IESKF until convergence is achieved. Finally, the optimal state is passed to the IMU model for accurate interpolation in subsequent scans. The data points are projected into ikd-Tree after assessing their uncertainty.}
	\label{fig:overview}
	\vspace{-5mm}
\end{figure}

Differing from the existing methods, this paper proposes an asynchronous multiple LiDAR-inertial odometry (\figref{fig:overview}). To deal with the abovementioned challenges in multi-LiDAR, we model point-wise uncertainty by considering the range and state covariance at each time. Furthermore, we calculate localization weight using the surrounding environment to determine the weight term during optimization. Handling a large number of points efficiently, we exploit ikd-Tree \cite{xu2022fast} and an \ac{IESKF}, achieving  computational efficiency. Our contributions are as follows:

\begin{enumerate}
    \item We tackle \ac{FOV} discrepancy by accurately transferring points among \ac{LiDAR}s, which the scan matching approach failed to solve. Furthermore, by utilizing \ac{CT} B-spline interpolation, we reduce temporal discrepancy, thus yielding consistency in inter-LiDAR scan alignment even with a significant FOV difference.
    \item The proposed point-level uncertainty captures increased ambiguity induced by range and point acquisition time. This point-wise assessment assigns larger uncertainty for points farther in range or later in time, hence handling uncertainty more generically.
    %\hl{residual vibration at longer ranges.}
    %We model point-wise uncertainty by considering the state covariance based on the IMU acquisition time, and 
    \item{The proposed localization weight balances the ratio between prior and measurement residual during optimization. This enables to automatic adapt the proportion of each residual in degenerate environments such as tunnels and narrow corridors.}
    \item Our method is validated on public and our own datasets. It is compatible with any combination of LiDAR with different scanning patterns from various manufacturers.
\end{enumerate}

%Exploiting direct approach \cite{chen2022direct}, the proposed method is applicable to various scanning pattern.

%FIGURE
%FIGURE
\section{METHOD}
\label{sec:method}

%----------------------------------------------------------------------%
\subsection{The Notion and State}

Subscript $A$ in notation $()_A$ denotes the representing frame. The frame $B$ in frame $A$ is denoted as $()_{AB}$. The ground truth is represented as $()$, while propagated, error, and optimal state are denoted as $\widehat{()}$, $\tilde{()}$, and $\bar{()}$.
For simplicity, we classify $N$ LiDARs as $\{L_i, i = 1, \cdots, N\}$, designating the LiDAR with the latest sample point as $P$ and all other LiDARs as $S$.
Our system comprises a state $\textbf{x}$, input $\textbf{u}$, and noise $\textbf{w}$ as

\vspace{-4mm}\small
\begin{eqnarray}
	\mathcal{M} &\triangleq& SO(3) \times \mathbb{R}^{15} \times \,\Pi_{i=1}^{N} \left(SO(3) \times \mathbb{R}^{3}\right) \nonumber \\
	    \textbf{x} &\triangleq& [\textbf{R}_{GI}^T \quad \textbf{t}_{GI}^T \quad \textbf{v}_{GI}^T \quad \textbf{b}_{\omega}^T \quad \textbf{b}_a^T \quad \textbf{g}_G^T \quad \textbf{R}_{IL_{i}}^T \quad   \textbf{t}_{IL_{i}}^T ]^T \in\mathcal{M} \nonumber \\
	\textbf{u} &\triangleq& \left[\omega_m^T \quad  a_m^T \right]^T,\, \textbf{w} \triangleq \left[n_\omega^T \quad n_a^T \quad n_{b\omega}^T \quad n_{ba}^T  \right]^T.
\end{eqnarray}
\normalsize
For the state $\textbf{x}$, transformation of the IMU frame (denoted as $I$) in the global frame (denoted as $G$) is $\textbf{T}_{GI} = \left(\textbf{R}^T_{GI}, \textbf{t}^T_{GI}\right)$, which consists of the rotation and translation. Additionally, \textbf{v}, \textbf{g}, and \textbf{b} stand for velocity, bias, and gravity, while $\textbf{T}_{IL}$ denote the extrinsic between LiDAR and IMU. $\{\omega_{m}, a_{m}\}$ denotes the angular velocity and linear acceleration from IMU sensor. Finally, \textbf{w} contains the white noise of these variables.

%----------------------------------------------------------------------%
\subsection{IMU Discrete Model with B-Spline Interpolation}
\label{sec:interoplation}

The \ac{CT} kinematic model can be converted into a discrete model using the $\boxplus$ outlined in \cite{xu2022fast} as

\vspace{-5mm}\small
\begin{eqnarray}
\label{eq:box}
	\textbf{x}_{i+1} &=& \textbf{x}_{i} \boxplus \left(\Delta \textnormal{t}\textbf{f}\left(\textbf{x}_i, \textbf{u}_i, \textbf{w}_i\right)\right), \\
	\textbf{f}\left(\textbf{x}, \textbf{u}, \textbf{w}\right) &=& \begin{bmatrix} \omega_m - b_\omega - n_\omega \\
	{\textbf{v}}_{GI} + \frac{1}{2}\left({\textbf{R}}_{GI} \left(a_m - b_a - n_a\right)+\textbf{g}_G\right)\Delta \textnormal{t} \nonumber \\
	{\textbf{R}}_{GI} \left(a_m - b_a - n_a\right)+\textbf{g}_G \\
	n_{b\omega} \\
	n_{ba} \\
	0_{3\times1} \\
	0_{3\times{2N}} \end{bmatrix}
\end{eqnarray}
\normalsize
Here, the function $\textbf{f}$ describes the state of the system to its subsequent state and is parameterized by the discretization interval $\Delta \textnormal{t}$. During the time interval between $(i-1)$ and $(i)^{\text{th}}$ scans, the system estimates the trajectory using the IMU, assuming the $(i-1)^{\text{th}}$ state to be optimal.
\begin{eqnarray}
	\label{eq:prop}
	\widehat{\textbf{x}}_{k+1} &=& \widehat{\textbf{x}}_k \boxplus \left(\Delta \textnormal{t} \textbf{f}\left(\widehat{\textbf{x}}_k, \textbf{u}_k, \textbf{0}\right)\right); \,
	\widehat{\textbf{x}}_0 = \bar{\textbf{x}}_{i-1}\\
	\label{eq:cov}
	\widehat{\Sigma}_{k+1} &=& \textbf{F}_{\tilde{\textbf{x}}_k}\widehat{\Sigma}_{k}\textbf{F}^T_{\tilde{\textbf{x}}_i} + \textbf{F}_{\textbf{w}_k}\textbf{Q}_{k}\textbf{F}^T_{{\textbf{w}}_i}; \, \widehat{\Sigma}_0 = \bar{\Sigma}_{i-1},
\end{eqnarray}
, where $Q_k$ denotes the covariance of $w_k$ while the jacobians $\textbf{F}_{\tilde{x}_k}$ and $\textbf{F}_{w_k}$ represent the derivatives of $\delta \left(\textbf{x}_{k+1} \boxminus \widehat{\textbf{x}}_{k+1}\right)$ with respect to each subscript, under the conditions that $(\tilde{\textbf{x}}_k, w_k) = (0, 0)$. Also, $\widehat{\textbf{x}}_{k-1}$ can be achieved using the $\boxminus$ operator as detailed in \cite{xu2021fast}. The accuracy of the interpolation is directly dependent on the accuracy of the discrete model. Since the IMU discrete model before optimization may not be accurate, the IMU state preceding $\bar{\textbf{x}}_{i-1}$ is recalculated via $\boxminus$ operator.

\begin{figure}[!t]
	\centering
		\includegraphics[width=1\columnwidth]{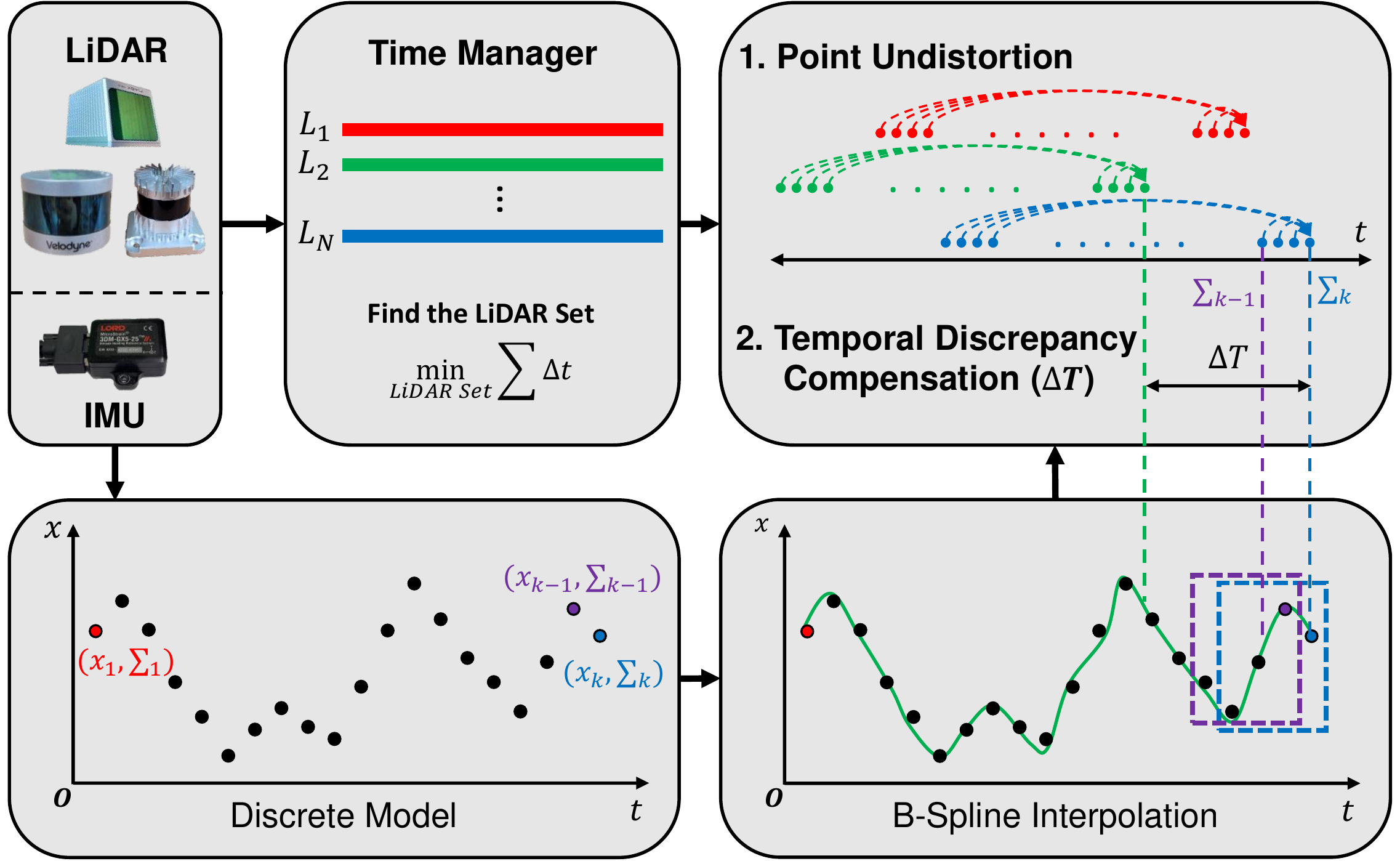}
  	\caption{IMU inputs are propagated by a discrete model and fed into B-spline interpolation. We search the LiDAR that satisfies the minimal time differences. Based on the interpolation, (\textit{i}) a pointcloud from LiDAR is undistorted by transformed into a single frame (the last point) and (\textit{ii}) the temporal discrepancy is compensated by the relative transformation among the last points in each LiDAR.}
	\label{fig:Method}
	\vspace{-6mm}
\end{figure}

Based on this propagation, B-spline interpolation is performed utilizing four transformations in the world frame, known as control points from $\textbf{T}^{k-1}_{GI}$ to $\textbf{T}^{k+2}_{GI}$ \cite{mueggler2018continuous}. By this interpolation, the trajectory at any time can be estimated, which is especially beneficial for asynchronous sensor systems. Furthermore, it is highly effective in environments with curved trajectories due to smoothness of spline. The system estimates the transformation at any time $t \in [t_k, t_{k+1})$ using
\begin{eqnarray}
  ^B\textbf{T}_{GI}\left(s\left(t\right)\right) = \textbf{T}^{k-1}_{GI} \Pi^{3}_{n=1}\exp\left(\tilde{\textbf{B}}_j\left(\s\left(t\right)\right)\Omega_{k+n-1}\right),
\end{eqnarray}
, while $t_k=k \Delta  t$ with $\Delta  t = t_{k+1} - t_{k}$, and $s(t) = \left(t - t_{k}\right) / \Delta  t$. $\Omega_k$ denotes the incremental control pose which is calculated as $\Omega_k = \log \left(\left(\textbf{T}^{k-1}_{GI}\right)^{-1}\textbf{T}^k_{GI}\right)$. $\tilde{\textbf{B}}_j$ is the matrix that includes the square term of $s$, and notation $^B\textbf{T}$ means that it is calculated by B-spline. Lastly, to obtain the covariance calculated in \eqref{eq:cov}, $\widehat{\Sigma}_{k+1}$ is assigned to the transformation achieved at $t \in [t_k, t_{k+1})$. The overall process is illustrated in \figref{fig:Method}.

%FIGURE
%FIGURE

%----------------------------------------------------------------------%
\subsection{Preprocessing of Multi-LiDAR System}

%Next, we need to merge pointclouds from multiple LiDAR.
Despite the strict interpolation, the interpolated pose accuracy is directly affected by the minimum difference in arrival times among LiDARs. Therefore, a set of LiDAR is chosen to minimize the sum of differences in the arrival times.

For undistortion, we focus on a single LiDAR, $S$, applying the same process to others. The first step of distortion starts with merging points obtained at different times into a single frame. This is achieved by determining the relative transformation between frames. Previous studies have utilized approximate discretized IMU or linear interpolation for the inference. Instead, by using B-spline based interpolation, the point $p_{S^j}$ at time $t_j$ can be transformed to the frame at $t_l$ to obtain undistorted point $p^u_{S^j}$ as
\begin{eqnarray}
\label{eq:distortion}
  p^u_{S^j} = \textbf{T}_{IS}^{-1} {}^B\textbf{T}_{{I^lI^j}} \textbf{T}_{IS} p_{S^j},
\end{eqnarray}
, where $t_l$ is the latest arrival time in $S$, and undistorted point is identified by notation $u$. Every LiDAR has different arrival times, and the temporal discrepancies should be compensated individually. 
When merging, the latest arrival time ($t_i$) of the latest LiDAR (P, denoted as blue point in \figref{fig:Method}) is leveraged to transform points acquired by other LiDARs.
%When merging, the points acquired by each LiDAR are transformed to the frame of reference corresponding to the final sampling time, $t_i$, of LiDAR $P$ (blue point in \figref{fig:Method}) with the latest time:
%
\begin{eqnarray}
\label{eq:temporal}
  p_{P^iS^j} = \textbf{T}_{IP}^{-1} {}^B\textbf{T}_{{I^iI^l}} \textbf{T}_{IS} p^u_{S^j}
= \textbf{T}_{IP}^{-1} {}^B\textbf{T}_{{I^iI^j}} \textbf{T}_{IS} p_{S^j}.
\end{eqnarray}
This transformation incorporates both undistortion and temporal compensation over multiple frame changes, and errors associated with these changes may be accumulated.

%----------------------------------------------------------------------%
\subsection{Uncertainty Propagation}

This uncertainty must be propagated to each point using the covariance to include the errors in the optimization. Covariances of ${^B}\textbf{T}$ and $\textbf{T}_{IL}$ are achieved by \eqref{eq:cov} and \ac{IESKF}. Also, the covariance of inverse transformation is calculated through $\Sigma_{inv} = \mathit{T} \Sigma \mathit{T}^T$, where $\mathit{T}$ is the adjoint matrix of $\textbf{T}^{-1}$. With fourth-order approximation, the transformation and covariance can be combined into $\{\textbf{T}_{{P^i}{S^j}}, \Sigma_{{P^i}{S^j}} \}$ \cite{barfoot2014associating}. The trace of $\Sigma_{{P^i}{S^j}}$ is referred to as the acquisition time uncertainty, which is visualized in \figref{fig:main}.

Three critical improvements beyond previous studies are as follows. Firstly, we distinguish the uncertainty according to the point sampling time, in contrast to assuming the same uncertainty in \cite{jiao2021robust}. Doing so allows a more accurate uncertainty modeling associated with each point.
Secondly, the specification of the primary sensor is no longer needed. In \cite{jiao2021robust}, because extrinsic covariance is only combined in the secondary LiDAR, the covariance of the secondary LiDAR is always higher than that of the primary LiDAR. Unlike theses, ours is more generic without specifying the primary explicitly. We utilize extrinsic covariances between LiDAR and IMU, resulting in all of the covariances being combined with the extrinsic covariance. This yields the covariances being equally affected by the extrinsic covariance. Finally, in contrast to the \cite{jiao2021robust}, which propagates uncertainty in the global frame, our approach propagates uncertainty on a per-point basis by confining it to the LiDAR frame. This decision aimed to account for the uncertainty introduced when fusing individual LiDAR points. By transforming a point into the frame $P$, the transformed point is presented as
%Since the point-wise uncertainty is used for evaluating the measurement uncertainty, we confined it to only the LiDAR frame with
%
\begin{eqnarray}
\label{eq:uncertain}
  p_{P^iS^j} &\triangleq& \widehat{\textbf{T}}_{{P^i}{S^j}}  \widehat{p}_{S^j} = \exp\left(\xi^{\wedge}_{{P^i}{S^j}}\right)\textbf{T}_{{P^i}{S^j}} \left({p}_{S^j} + D\zeta\right)\\
  &\approx& \left(I + \xi^{\wedge}_{{P^i}{S^j}}\right)\textbf{T}_{{P^i}{S^j}} \left({p}_{S^j} + D\zeta\right) \nonumber
\end{eqnarray}
Here, $\xi$ is the error of the transformation and $\zeta \in \mathbb{R}^3$ is the perturbation of the LiDAR measurement. Also, $p$ in \eqref{eq:uncertain} is a $4\times1$ vector with a scale value $1$ added, and $D$ is the dilation matrix which transforms the dimension from $3 \times 1$ to $4 \times 1$, with zero terms added. As the second-order error is computationally expensive but has a relatively small effect, we only consider the first-order term, $\widehat{\textbf{T}}\widehat{p} \approx \textbf{q} + \textbf{Q}\theta$.

\vspace{-3mm}\small
\begin{eqnarray}
\label{eq:uncertain2}
  &\textbf{q}& := \textbf{T}p, \  \textbf{Q} := \left[ \left(\textbf{T}p\right)^{\odot} \  \textbf{T}D\right], \  \begin{bmatrix} \epsilon \\ \eta\end{bmatrix}^{\odot} := \begin{bmatrix}\eta\textbf{1} \quad - \epsilon^{\wedge} \\ \textbf{0}^{T} \quad \quad \textbf{0}^{T} \end{bmatrix} \\
  &\theta& := \left[\xi^T \quad \zeta^T\right]^T, \  \theta \sim \mathcal{N}\left(\textbf{0}, \Xi\right), \  \Xi = \textnormal{diag}\left(\Sigma_{P^iS^j}, \textbf{Z}\right), \nonumber
\end{eqnarray}
\normalsize
with the LiDAR measurement covariance $\textbf{Z}$. Since $p_{P^iS^j}$ follows the Gaussian distribution, the uncertainty of the point can be obtained through \eqref{eq:uncertain2}, as $\Sigma_p = Q \Xi Q^T$. The resulting uncertainty is derived from the range of point $p$ from $\left(\textbf{T}p\right)^{\odot}$, and the acquisition time from $\Sigma_{P^iS^j}$. Improved accuracy is achieved by addressing ambiguity from IMU discrete model error over time and noise effects which increase with range from vibration. It is utilized for the optimization with $p_{S^j}$.

%----------------------------------------------------------------------%
\subsection{Measurement Model with Uncertainty}

Our system utilizes a surface measurement model without feature extraction under the assumption of local planarity. The points, $\{\textbf{p}_{S^j}, j = 1, \cdots , l\}$ in the LiDAR frame can be transformed to the world frame with the following equation:
\begin{eqnarray}
  \textbf{p}_{GS^j} = \textbf{T}_{GI^i}\textbf{T}_{IP} \textbf{p}_{P^iS^j} = \textbf{T}_{GI^i} {}^B\textbf{T}_{{I^iI^l}} \textbf{T}_{IS} \textbf{p}^u_{S^j}
\end{eqnarray}
For the lastly sampled point LiDAR $P$, ${}^B\textbf{T}_{{GI}^i}^{-1} {}^B\textbf{T}_{{GI}^l}$ is equivalent to the identity matrix as there is no temporal discrepancy to be compensated. To approximate the surface, our method selects the five nearest neighbor points from the measurement in the ikd-Tree. In doing so, our system incorporates the associated uncertainty of points in the ikd-Tree into the measurement model. For generality, we use the notation $L$, indicating the LiDAR, $S$, or $P$. The weighted sample covariance of the plane for point $p_{L^j}$, $\Sigma_{L^j}$ is calculated by
\begin{eqnarray}
  \Sigma_{L^j} = \sum_{n=1}^{5}w^2_n \Sigma_n, \quad w_n = \frac{\tau - \textnormal{tr}\left(\Sigma_n\right)}{\sum_{n=1}^{5}[\tau-\textnormal{tr}\left(\Sigma_n\right)]}
\end{eqnarray}
, while $\tau$ represents the uncertainty threshold, which is also utilized in the mapping process in the subsequent section. Based on the normal vector of the plane, $\textbf{v}_{GL^j}$, and the plane covariance, $\Sigma_p$, the measurement model is calculated as
\begin{eqnarray}
\label{eq:modelFull}
  \textbf{0} = \textbf{h}_{L^j}\left(\textbf{x}_i, \, \textbf{n}_{L^j}\right) = \frac{\textbf{v}^T_{GL^j}\left(\textbf{T}_{GI^i} {}^B\textbf{T}_{{I^iI^l}} \textbf{T}_{IL} \left(\textbf{p}^u_{L^j}+ \textbf{n}_{L^j}\right) - \textbf{q}_{GL^j}\right)}{\textnormal{FIC}\left(\textnormal{tr}\left(\Sigma_{L^j}\right), s_{max}, s_{min}\right)}
\end{eqnarray}
Here, $\textbf{n}_{L^j}$ represents the noise from the LiDAR, and $\textbf{q}_{GL^j}$ is a point located on the plane. Additionally, $\textbf{h}$ represents the measurement model, which is summary of the terms of state, including $\textbf{T}_{GI}$ and $\textbf{T}_{IL}$. We employ \ac{FIC} to bind the uncertainty, which is calculated as
\begin{eqnarray}
\label{eq:FIC}
  \textnormal{FIC}(V, I_{max}, I_{min}) = \frac{(I_{max} - I_{min})(V-V_{min})}{V_{max}-V_{min}} + I_{min},
\end{eqnarray}
with $I_{max}$ and $I_{min}$ to be the rescaling interval. Similarly, $V_{max}$ and $V_{min}$ are the maximum and minimum values in $V$. Utilizing FIC, we balance measurement influence by adjusting the values within set bounds, which regulates performance and reliability without overemphasis or neglect.

%----------------------------------------------------------------------%
\subsection{Iterated Error State Kalman Filter}

Our state estimation comprises three components as in \figref{fig:overview}, namely state propagation, residual calculation, and state update. The state propagation component, as represented in equations \eqref{eq:prop} and \eqref{eq:cov}, is utilized as the prior distribution, and its error state is obtained through
\begin{eqnarray}
\label{eq:distribution1}
\textbf{x}_i \boxminus \widehat{\textbf{x}}_i = \left(\widehat{\textbf{x}}_i^{\kappa} \boxplus \tilde{\textbf{x}}_i^{\kappa}\right) \boxminus \widehat{\textbf{x}}_i = \widehat{\textbf{x}}_i^{\kappa} \boxminus \widehat{\textbf{x}}_i + \textbf{J}^{\kappa} \tilde{\textbf{x}}_i^{\kappa} \sim \mathcal{N}(\textbf{0}, \widehat{\Sigma}_i).
\end{eqnarray}
$\textbf{J}^{\kappa}$ represents the Jacobian matrix of  $\left(\widehat{\textbf{x}}_i^{\kappa} \boxplus \tilde{\textbf{x}}_i^{\kappa}\right) \boxminus \widehat{\textbf{x}}_i$ with the condition that $\tilde{\textbf{x}}_i^{\kappa} = 0$. When $\kappa = 1$, $\textbf{J}^{\kappa} = \textbf{I}$ and transformation term in $\widehat{\textbf{x}}_{i}$ becomes ${}^B\textbf{T}_{GI^i}$. Further details are as in \cite{xu2021fast}.

In the case of the measurement model, another distribution can be identified through a first-order approximation:
\begin{eqnarray}
\label{eq:distribution2}
  \textbf{0} &=& \textbf{h}_{L^j}\left(\textbf{x}_i, \textbf{n}_{L^j}\right) \simeq \textbf{h}_{L^j}\left(\widehat{\textbf{x}}_i^{\kappa}, \textbf{0}\right) + \textbf{H}^{\kappa}_{L^j} \tilde{\textbf{x}}^{\kappa}_k + \textbf{v}_{L^j} \\
  -\textbf{v}_j &=& \textbf{z}^{\kappa}_{L^j} + \textbf{H}^{\kappa}_{L^j} \tilde{\textbf{x}}^{\kappa}_k \sim \mathcal{N}\left(\textbf{0}, \Sigma_{L^j}\right) \nonumber
\end{eqnarray}
, where $\textbf{H}^{\kappa}_{L^j}$ is the Jacobian of $\textbf{h}_{L^j}\left(\widehat{\textbf{x}}_i^{\kappa} \boxplus \tilde{\textbf{x}}_i^{\kappa}, \textbf{n}_{L^j}\right)$ with $\tilde{\textbf{x}}^{\kappa}_i$, and $\textbf{v}_j$ is the noise with covariance calculated as in \eqref{eq:uncertain2}.

Utilizing both prior distribution \eqref{eq:distribution1} and measurement distribution \eqref{eq:distribution2}, the state estimation problem is changed into \ac{MAP}:
\begin{eqnarray}
  \min\limits_{\tilde{\textbf{x}}^{\kappa}_i} \left( \lVert \textbf{x}_i \boxminus \widehat{\textbf{x}}_i \rVert ^2_{\widehat{\Sigma}_i} + w_l^2\sum_{L = P, S}\sum^{m}_{j=1} \lVert \textbf{z}^{\kappa}_{L^j} + \textbf{H}^{\kappa}_{L^j} \tilde{\textbf{x}}^{\kappa}_{i} \rVert ^{2}_{\textbf{R}_{L^j}} \right)
\end{eqnarray}
, while $\textbf{R}_{L^j}$ is the output from $\textnormal{FIC}(\textnormal{tr}(\Sigma_{L^j}), \textbf{R}_{max}, \textbf{R}_{min})$, and $\lVert \textbf{x}\rVert^2_{\Sigma} = \textbf{x}^T\Sigma^{-1} \textbf{x}$. The localization weight, $w_l$, is given to the prior distribution over measurement, especially for the degenerated environment. The value of $w_l$ can be determined by taking the ratio of $\sigma_1$ to $\sigma_3$, which are obtained from \ac{SVD} of the normal vector of the measurement.
If $w$ falls outside of the boundary of $({b}_{min}, {b}_{max})$, the values of $l_{min}$ and $l_{max}$ are assigned to $w_l$, respectively. Otherwise, the value of $w_l$ can be obtained as $w_l = \textnormal{FIC}(w, l_{max}, l_{min})$. An iterated Kalman filter can solve this maximum a posteriori (MAP) problem.

\vspace{-5mm}\small
\begin{eqnarray}
  \textbf{K} &=& { \left( \textbf{H}^T \textbf{R}^{-1} \textbf{H} + \textbf{P} ^{-1} \right) }^{-1} \textbf{H}^T \textbf{R}^{-1} ,\\
  {\widehat{\textbf{x}}}^{\kappa + 1}_i &=& {\widehat{\textbf{x}}}^{\kappa}_i \boxplus \left(   - \textbf{K} \textbf{z}^{\kappa}_i - \left( \textbf{I} - \textbf{KH} \right) {\left(  \textbf{J}^{\kappa}  \right)}^{-1}  \left(  {\widehat{\textbf{x}}}^{\kappa}_{i} \boxminus  {\widehat{\textbf{x}}}_{i}  \right)  \right) \nonumber
\end{eqnarray}
\normalsize
with $\small \textbf{H} = w_l \times {\left[  {\textbf{H}^{\kappa}_{S^1}}^T, \cdots, {\textbf{H}^{\kappa}_{S^m}}^T, {\textbf{H}^{\kappa}_{P^1}}^T, \cdots, {\textbf{H}^{\kappa}_{P^m}}^T   \right]}^T  $,
$\small \textbf{R} = \text{diag} \left(  \textbf{R}_{S^1}, \cdots,  \textbf{R}_{S^m}, \textbf{R}_{P^1}, \cdots, \textbf{R}_{P^m}  \right)  $,
$\small \textbf{P} = \left( \textbf{J}^{\kappa} \right)^{-1} \widehat{\Sigma}_i   \left( \textbf{J}^{\kappa} \right)^{-T} $,
and $\small \textbf{z}^{\kappa}_{i} = w_l \times {\left[  {\textbf{z}^{\kappa}_{S^1}}^T , \cdots ,  {\textbf{z}^{\kappa}_{S^m}}^T, {\textbf{z}^{\kappa}_{P^1}}^T , \cdots ,  {\textbf{z}^{\kappa}_{P^m}}^T   \right]}^T $.
The iterative process is repeated until the convergence criteria, which $\lVert \widehat{\textbf{x}}_i^{\kappa+1} \boxminus \widehat{\textbf{x}}_i^{\kappa}\rVert < \epsilon$ is satisfied. The final estimates of the state and its corresponding covariance are as:
%Thanks to the previous work \cite{xu2021fast} on the Kalman gain, the dimensionality of the calculation has been reduced to the state dimension.
%
\begin{eqnarray}
  \bar{\textbf{x}}_i = \widehat{\textbf{x}}_i^{\kappa+1}, \quad \bar{\Sigma} = \left( \textbf{I} -\textbf{K}\textbf{H} \right)\textbf{P}
\end{eqnarray}

%----------------------------------------------------------------------%
\subsection{Mapping with Uncertainty}

Using the estimated state, $\bar{\textbf{x}}_i$, the pointcloud from LiDAR can be transformed into the world frame,
\begin{eqnarray}
  \bar{\textbf{p}}_{L^j} = \bar{\textbf{T}}_{GI^i} {}^B\textbf{T}_{{I^iI^l}} \bar{\textbf{T}}_{IL} \textbf{p}^u_{L^j}.
\end{eqnarray}

To effectively maintain the ikd-Tree, the uncertainty of the point is first evaluated. If the uncertainty of a point, $\textnormal{tr}(\Sigma_{L^j})$, is larger than a predefined $\tau$, that point is not saved in the tree. The saved points are then added to the tree via downsampling. In contrast to the original ikd-Tree, our proposed strategy considers uncertainty during the insertion process. The downsampling process is designed to retain only points close to the center of the tree resolution for accurate mapping. In our implementation, if the insertion point lies within the diagonal value of \textbf{Z} at the center of the tree, points with low uncertainty are retained in the tree.

\section{experiment}
\label{sec:experiment}

%TABLE
\begin{table}[t] 
\centering
\caption{Dataset description}
\label{tab:dataset}
\resizebox{\columnwidth}{!}{\begin{tabular}{c|c|c|c|c}
\toprule
Dataset                   & Number & LiDAR         & IMU                    & Environment                             \\ \midrule
\multirow{2}{*}{Hilti}    & 1      & OS0-64        & \multirow{2}{*}{200Hz} & \multirow{2}{*}{Indoor \& Outdoor}      \\
                          & 2      & Livox Horizon &                        &                                         \\ \midrule
\multirow{3}{*}{UrbanNav} & 1      & HDL-32E       & \multirow{3}{*}{400Hz} & \multirow{3}{*}{Urban with Skyscrapper} \\
                          & 2      & VLP-16        &                        &                                         \\
                          & 3      & LS-16C        &                        &                                         \\ \midrule
\multirow{3}{*}{Ours}     & 1      & OS2-128       & \multirow{3}{*}{100Hz} & \multirow{3}{*}{Urban with Challenges}  \\
                          & 2      & Livox Avia    &                        &                                         \\
                          & 3      & Livox Tele    &                        &           \\ \bottomrule                             
\end{tabular}}
\vspace{-4mm}
\end{table}
%TABLE
%TABLE
\begin{table}[!t]
\vspace{2mm}
\caption{ATE$_t$ for Hilti SLAM Dataset 2021}
\label{tab:Hilti}
\centering
\begin{tabular}{cccccc} \toprule
             & Ours  & Fast-H & Fast-O    & M-LOAM & LOCUS 2.0         \\ \midrule
\texttt{Basement}     & \textbf{0.036} & 0.709        & \textit{0.046} & 0.115  & 0.120          \\
\texttt{Campus}       & \textbf{0.046} & 0.063        & \textit{0.063} & 0.386  & 0.087          \\
\texttt{Construct} & \textbf{0.063} & 0.200        & \textit{0.088} & 2.647  & 0.290          \\
\texttt{LAB}          & \textbf{0.024} & Err          & \textit{0.026} & 0.064  & 0.040          \\
\texttt{UZH}          & \textbf{0.177} & 0.233        & 0.184          & 0.276  & \textit{0.177} \\ \bottomrule
\end{tabular}

\footnotesize
\raggedright
\vspace{0.8mm}
The best results are in \textbf{bold} and the second-best's are in \textit{italic}.
\vspace{-5mm}
\end{table}

%TABLE
%----------------------------------------------------------------------%
\subsection{Dataset and Evaluation}
\label{sec:Dataset}

\subsubsection{Dataset}

To evaluate the performance of our method in various environments, we conduct experiments on three datasets: Hilti SLAM Dataset 2021 \cite{helmberger2022hilti}, UrbanNav \cite{hsu2021urbannav}, and our dataset. In the Hilti dataset, we evaluate our method for a hand-held system in small-scale indoor and outdoor environments. The UrbanNav dataset is exploited to assess the performance of a method for a vehicle-like system in an urban environment at a large scale. In addition to public datasets, we collect our dataset to assess the performance in challenging environments at a higher speed ($\sim$50 \unit{km/h}) including U-turns and tunnels. {To synchronize the temporal foundation of each sensor, \ac{PTP} was employed. However, this does not imply that all sensors were firing simultaneously.} The datasets are listed in \tabref{tab:dataset}, with detailed descriptions of each sequence provided in subsequent sections.

\subsubsection{Evaluation}

Ours is compared with three state-of-the-art methods including {Fast-LIO2 (single) \cite{xu2022fast}, M-LOAM (multi) \cite{jiao2021robust}, and LOCUS 2.0 (multi) \cite{reinke2022locus}}. To ensure the fair comparisons, we employ the following strategies. Fast-LIO2 only supports single LiDAR, and we utilize the central LiDAR as it provides the most points. For the Hilti dataset, we obtain the trajectory from each LiDAR, OS0-64 (denoted as FAST-O) and Livox Horizon (denoted as Fast-H). For M-LOAM, which supports spinning LiDAR, we incorporate the points from Livox LiDAR as a surface feature. For LOCUS 2.0, we readjust the parameters for GICP to adapt to the specific environment. Notably, no odometry input is provided in the dataset. In our method, we select the parameters as {$\textbf{Z} = \textnormal{diag}(0.05, 0.05, 0.05)$}, $(s_{min}, s_{max}) = (1, 1.25)$, $(b_{min}, b_{max}) = (0.2, 0.8)$, $(l_{min}, l_{max}) = (0.5, 3)$, $(R_{min}, R_{max}) = (0.0075, 0.0125)$ and $\tau = 1$ with little variation.

To quantitatively compare the performance of methods, we calculate the \ac{RMSE} of the \ac{ATE} and \ac{RTE} using Evo evaluator \cite{grupp2017evo}. The \ac{ATE} is measured in meters and degrees, while the \ac{RTE} is measured in percentage and degree per meter. For the Hilti dataset, we leverage the evaluator provided by the dataset to calculate the ATE$_t$, translation component of the \ac{ATE}. While for our dataset, the ground truth is obtained using an \ac{INS}. To ensure the reliability of the results, we only utilize positions with the status, \texttt{INS\_SOLUTION\_FREE}.

%----------------------------------------------------------------------%
\subsection{Hilti SLAM Dataset 2021}

The results from the Hilti evaluator are presented in \tabref{tab:Hilti}. Among all methods, ours yields the most accurate results in all sequences. Interestingly, Fast-O achieves the second-best performance in most sequences, even without using multi-LiDAR. As expected, M-LOAM and LOCUS 2.0 are less accurate than the others, attributed to the lack of synchronization of inter LiDAR. In the case of \texttt{LAB} and \texttt{UZH}, the impact of asynchrony is minimal since intense movement does not exist in the indoor environment. However, in the case of \texttt{Campus} and \texttt{Construct}, which involve more vigorous movement, the errors are more pronounced due to the inherent asynchrony of the sensors. Despite these challenges, our method exhibits robustness in all sequences owing to its temporal compensation.

%FIGURE
\begin{figure}[t]
	\centering
		\includegraphics[width=0.97\columnwidth]{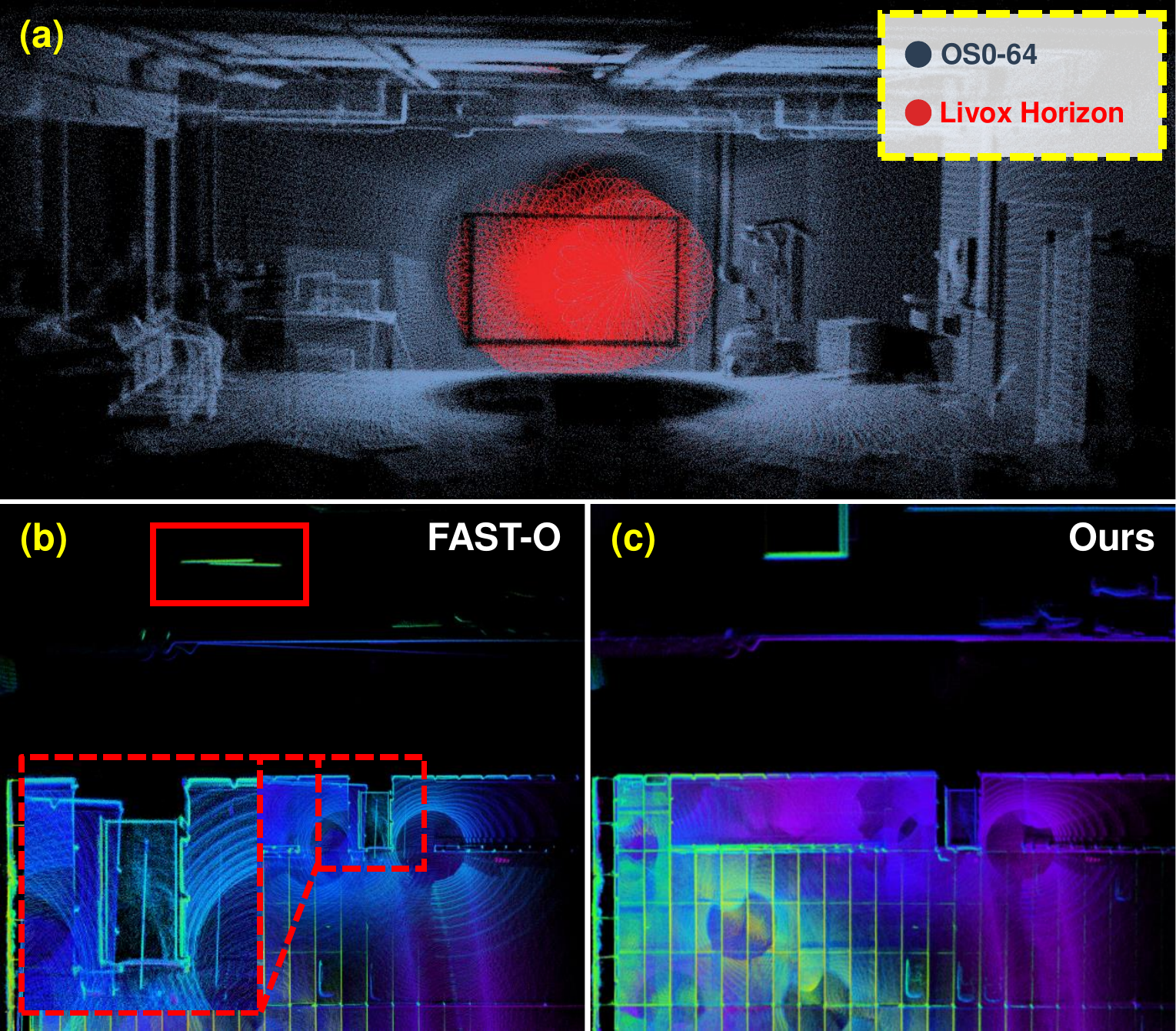}
  	\caption{(a) Accumulated scans from \texttt{LAB}. Compared to Ouster (gray), Livox (red) reveals very limited \ac{FOV} and causes localization failure due to the lack of distinct features. (b-c) show the outputs of Fast-O and our method for the \texttt{Parking} dataset. As shown in the red box in (b), the map produced by Fast-O is misaligned when returning to the starting point, while our result in (c) is well aligned.}
	\label{fig:Hilti}
	\vspace{-7mm}
\end{figure}
%FIGURE

Meanwhile, unlike Fast-O, Fast-H is degenerated significantly due to the limited \ac{FOV} of the LiDAR. The limited \ac{FOV} of Livox is prone to encounter degenerate cases as depicted in \figref{fig:Hilti}(a). This FOV-induced limitation may yield a substantial tracking deviation even in the presence of a rich feature. In this perspective, Fast-O seems the best choice for most environments, but it is not optimal in all cases, particularly in the \texttt{Parking}. Since the ground truth of \texttt{Parking} is not provided, we examine the reconstructed map for qualitative evaluation as illustrated in \figref{fig:Hilti}(b). In the center of the \texttt{Parking}, a substantial number of points are not detected, causing a tracking error and drift in the map. On the other hand, our method can mitigate these issues by incorporating additional LiDAR as shown in \figref{fig:Hilti}(c), demonstrating the robustness of the multi-LiDAR system.

%----------------------------------------------------------------------%
\subsection{UrbanNav Dataset}

The UrbanNav Dataset consists of three spinning LiDARs with two mounted on both sides in an inclined configuration. This configuration maximizes the scanning area but with minimal overlap, and can detect objects at higher elevations that are invisible to the central LiDAR. However, since this dataset contains multiple dynamic objects at a higher driving speed, inter-LiDAR transformation encompasses significant variations due to temporal discrepancies from asynchrony.

%TABLE

% Please add the following required packages to your document preamble:
% \usepackage{multirow}
\begin{table}[t]
    \scriptsize
    \caption{UrbanNav Dataset Evaluation}
    \label{table:urbanNav}
    \centering
    \resizebox{0.45\textwidth}{!}{\begin{tabular}{cccccc}
        \toprule
                                                &                             & \begin{tabular}[c]{@{}c@{}}Fast-LIO2\end{tabular}  & \begin{tabular}[c]{@{}c@{}}M-LOAM\end{tabular} & {LOCUS 2.0} & Ours          \\ \midrule[1pt]
    \multicolumn{1}{c|}{\multirow{4}{*}{\rotatebox{90}{\texttt{{Mongok}}}}} & \multicolumn{1}{c|}{ATE$_t$} & \textit{5.917}  & 25.899 & 6.846  & \textbf{2.579}                                       \\ 
    \multicolumn{1}{c|}{}                       & \multicolumn{1}{c|}{ATE$_r$} & \textit{4.039} & 9.140  & 5.616  & \textbf{2.383}                                             \\
    \multicolumn{1}{c|}{}                       & \multicolumn{1}{c|}{RTE$_t$} & 0.188   & 0.632 & \textit{0.174}  & {\textbf{0.167}}                                               \\
    \multicolumn{1}{c|}{}                       & \multicolumn{1}{c|}{RTE$_r$} & 0.749   & 1.006  & \textbf{0.710}  & \textit{0.736}                                                \\ \midrule
    \multicolumn{1}{c|}{\multirow{4}{*}{\rotatebox{90}{\texttt{{Whampoa}}}}} & \multicolumn{1}{c|}{ATE$_t$} & \textit{7.066}  & 31.482 & 18.124  &   \textbf{4.236}                                       \\
    \multicolumn{1}{c|}{}                       & \multicolumn{1}{c|}{ATE$_r$} & \textit{7.066}  & 8.286& 9.404  &   \textbf{4.600}                                   \\
    \multicolumn{1}{c|}{}                       & \multicolumn{1}{c|}{RTE$_t$} &  0.390  & 0.710& \textit{0.339} & {\textbf{0.207}}                                       \\
    \multicolumn{1}{c|}{}                       & \multicolumn{1}{c|}{RTE$_r$} & \textit{1.034}  & 1.213 & 1.238 & {\textbf{1.033}}                            \\ \midrule
    \multicolumn{1}{c|}{\multirow{4}{*}{\rotatebox{90}{\texttt{{TST}}}}} & \multicolumn{1}{c|}{ATE$_t$} & \textit{8.783}  & 53.682 & 33.292 & {\textbf{2.342}}                                       \\
    \multicolumn{1}{c|}{}                       & \multicolumn{1}{c|}{ATE$_r$} & \textit{6.640}  & 21.584& 13.367 & {\textbf{5.085}}                                    \\
    \multicolumn{1}{c|}{}                       & \multicolumn{1}{c|}{RTE$_t$} & \textit{0.494}  &  2.177 & 0.841 & \textbf{0.351}                                     \\
    \multicolumn{1}{c|}{}                       & \multicolumn{1}{c|}{RTE$_r$} & \textit{1.264}  & 1.355& 1.748 &   {\textbf{1.261}}                                      \\ 
    \bottomrule
    \end{tabular}}
    \vspace{-3mm}
        \end{table}

%TABLE

%FIGURE
\begin{figure}[!t]
	\centering
    \includegraphics[width=1\columnwidth]{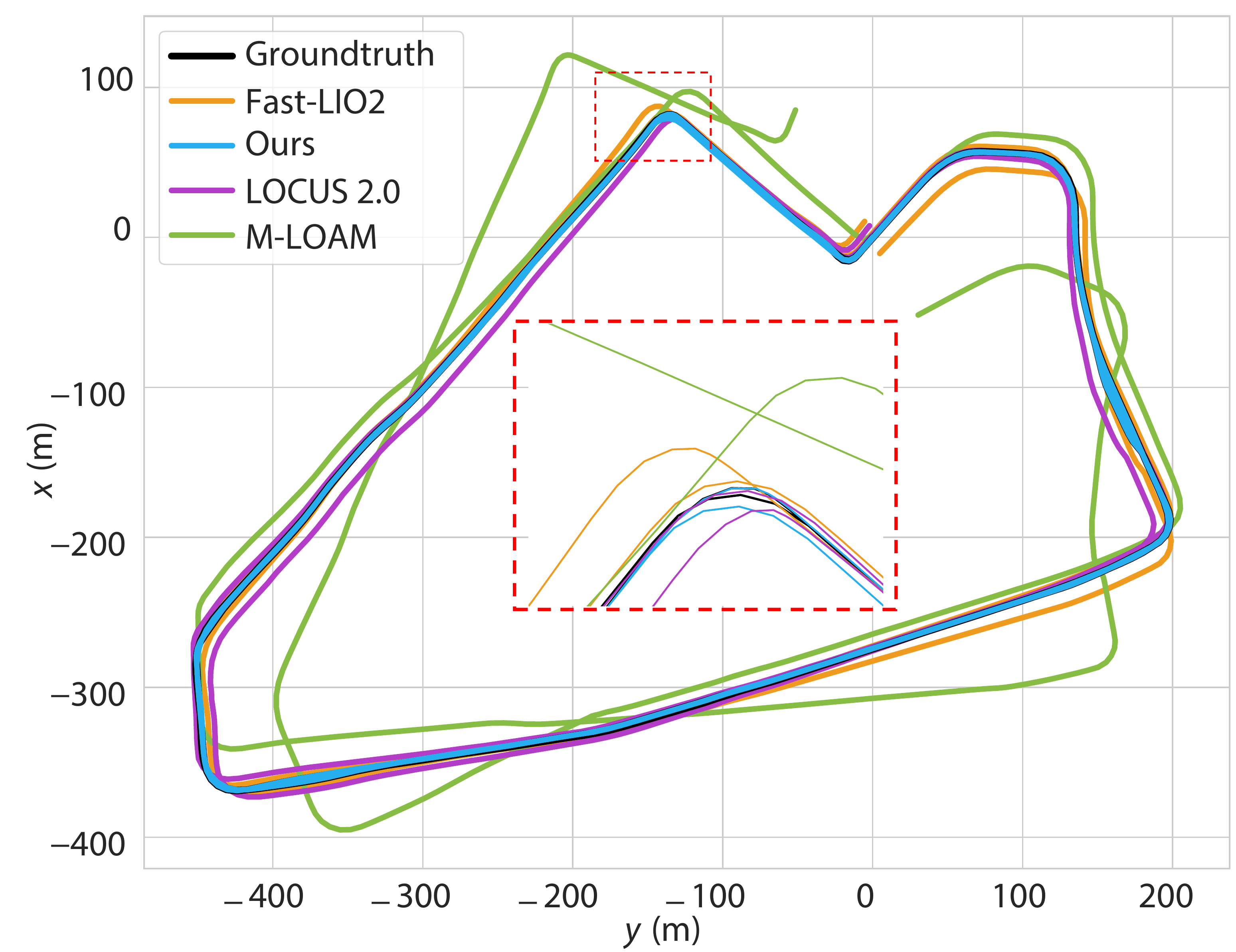}
  	\caption{\texttt{TST} results. Our method (blue) demonstrates a highly close alignment with the ground truth (black).}
	\label{fig:Urban}
	\vspace{-6mm}
\end{figure}
%FIGURE

As shown in Table \ref{table:urbanNav}, our method outperforms all others. In the \texttt{Mongok}, most methods show minor errors due to the repeated traversal of a single loop. However, \texttt{Whampoa} exhibits more significant error in most algorithms as there is an underpass in the middle of the trajectory (\figref{fig:tunnel}(b)). Additionally, the higher vehicle speed in the \texttt{TST}---approximately twice that of the Mongok scenario---has exacerbated the error of the multi-LiDAR system due to the difficulties in merging pointclouds in asynchronous systems. Despite these challenges, the proposed method can effectively suppress significant errors by accurately estimating the relative transformation inter-LiDAR and allowing extensive scanning, which is impossible with a single LiDAR (\figref{fig:Urban}).
%From the advantages, only a z-axis error exists, while the other axis showed high accuracy as illustrated in

%----------------------------------------------------------------------%
\subsection{Our Own Dataset}
\label{sec:Private}

%TABLE

% Please add the following required packages to your document preamble:
% \usepackage{multirow}
\begin{table}[t]
    \scriptsize
    \caption{Private Dataset Evaluation}
    \label{tab:Ourown}
    \centering
    \resizebox{0.45\textwidth}{!}{\begin{tabular}{ccccccc}
        \toprule
                                                &                             & \begin{tabular}[c]{@{}c@{}}Fast-LIO2\end{tabular}  & \begin{tabular}[c]{@{}c@{}}M-LOAM\end{tabular} & {LOCUS 2.0} & Ours          \\ \midrule[1pt]
    
    \multicolumn{1}{c|}{\multirow{4}{*}{\rotatebox{90}{\texttt{{City01}}}}} & \multicolumn{1}{c|}{ATE$_t$} & \textit{9.970}  & 33.907 & 23.998  & \textbf{6.538}                                    \\
    \multicolumn{1}{c|}{}                       & \multicolumn{1}{c|}{ATE$_r$} & \textit{4.575}  & 8.792& 5.521  & \textbf{3.491}                                       \\
    \multicolumn{1}{c|}{}                       & \multicolumn{1}{c|}{RTE$_t$} &  \textit{0.292}  & 0.955& 0.609 & \textbf{0.266}                                        \\
    \multicolumn{1}{c|}{}                       & \multicolumn{1}{c|}{RTE$_r$} & \textit{0.898}  & 1.020 & 0.895 & \textbf{0.874}                                      \\ \midrule
    \multicolumn{1}{c|}{\multirow{4}{*}{\rotatebox{90}{\texttt{{City02}}}}} & \multicolumn{1}{c|}{ATE$_t$} & \textit{35.308}  & 72.382 & 58.211 & \textbf{\textit{6.707}}                                     \\
    \multicolumn{1}{c|}{}                       & \multicolumn{1}{c|}{ATE$_r$} & 7.473  & \textit{4.683} & 4.722 & {\textbf{3.522}}                                     \\
    \multicolumn{1}{c|}{}                       & \multicolumn{1}{c|}{RTE$_t$} & \textit{0.608}  &  3.665 & 1.531 & \textbf{0.565}                                     \\
    \multicolumn{1}{c|}{}                       & \multicolumn{1}{c|}{RTE$_r$} & 1.179  & \textit{1.104}& 1.167 &   \textbf{1.084}                                       \\ 
     \midrule
    \multicolumn{1}{c|}{\multirow{4}{*}{\rotatebox{90}{\texttt{{City03}}}}} & \multicolumn{1}{c|}{ATE$_t$} & \textit{6.951}  & 33.801 & 21.753  & \textbf{5.470}                                            \\ 
    \multicolumn{1}{c|}{}                       & \multicolumn{1}{c|}{ATE$_r$} & \textit{4.194} & 6.657  & 4.773  & \textbf{3.522}                                           \\
    \multicolumn{1}{c|}{}                       & \multicolumn{1}{c|}{RTE$_t$} & \textit{0.996}   & 1.310 & 1.159  & \textbf{0.565}                                                \\
    \multicolumn{1}{c|}{}                       & \multicolumn{1}{c|}{RTE$_r$} & 1.088   & \textbf{1.070}  & 1.089  & \textit{1.084}                                                 \\ 
    \bottomrule
    \end{tabular}}
    \vspace{-3mm}
    \end{table}

%TABLE

%FIGURE
\begin{figure}[t]
	\centering
		\includegraphics[width=1\columnwidth]{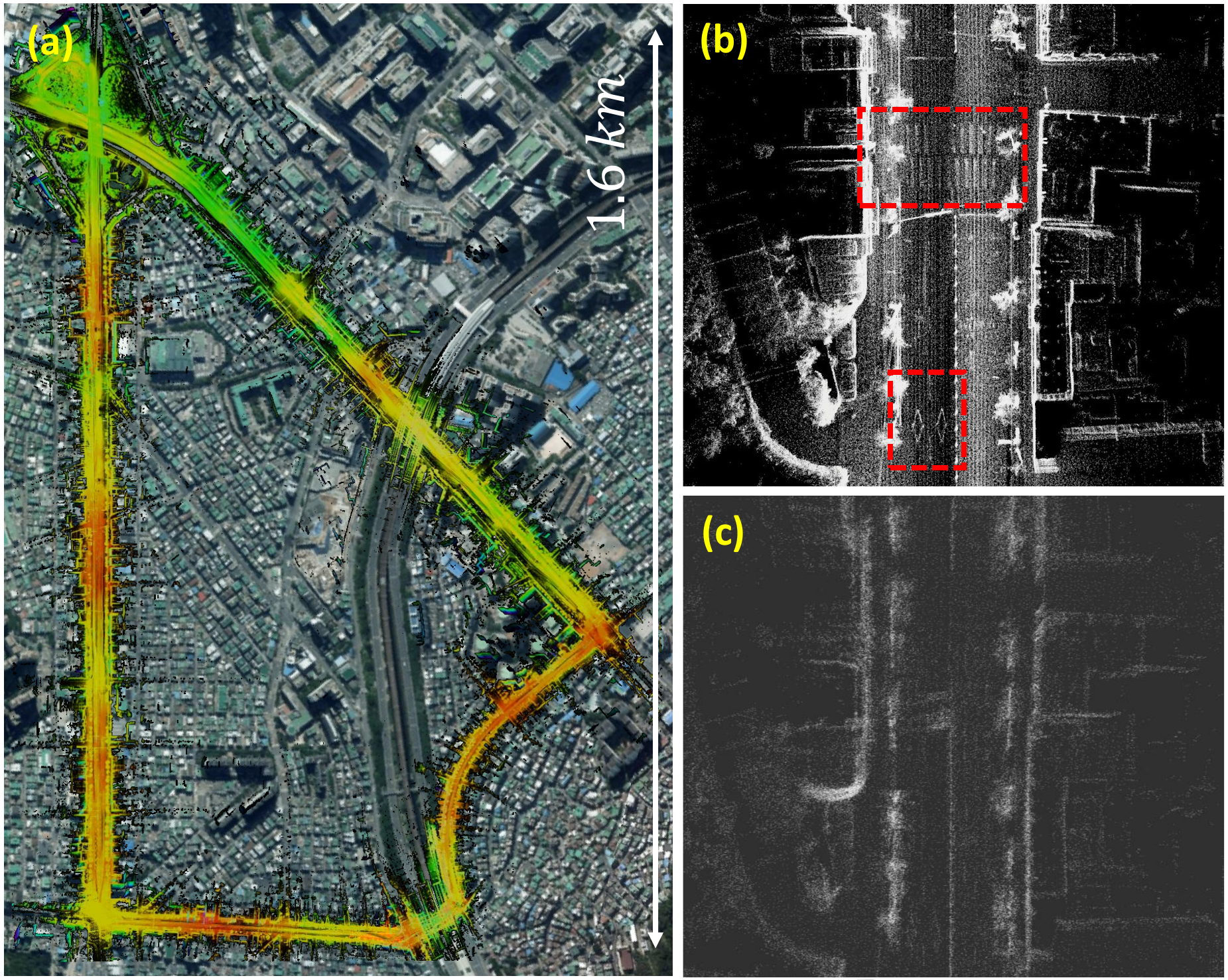}
  	\caption{(a) The resulting map of \texttt{City03}. Points are color-coded by their height from low (red) to high (green). On the right, scans of stacks are presented from (b) our method and (c) LOCUS 2.0. Our method effectively compensates for temporal discrepancies, resulting in a low noise level even when markers on the street are present, as in the red box.}
	\label{fig:Private}
	\vspace{-6mm}
\end{figure}
%FIGURE

Our dataset, \texttt{City01-03}, presents a unique set of challenges, whereas our proposed method achieves superior performance in most cases, as presented in \tabref{tab:Ourown}. \texttt{City01} includes many rotations and U-turns. After completing a U-turn, both M-LOAM and LOCUS 2.0 experienced localization failure when attempting to match the previously generated map. This failure is attributed to the lack of an initial estimation in LOCUS 2.0 and the inability to align the scans among LiDARs. As a result, the system resorts to constructing an additional map for localization.

\texttt{City02} features a tunnel environment with a length of approximately \unit{400}{m}. In this sequence, failure to establish correspondences between points in the tunnel interrupts estimating forward motion yielding a significant error.

Lastly, \texttt{City03} is over \unit{4.3}{km} with numerous dynamic objects and no loops until the return to the start point with a large accumulated error. Among others, our method exhibits low error in \texttt{City03}, even in the absence of a loop in the middle of the trajectory. For example, the map is well-aligned after completing a single lap as presented in \figref{fig:Private}. Despite using asynchronous LiDAR, our method can generate clarified maps with reduced noise thanks to our temporal compensation and accurate odometry.

%----------------------------------------------------------------------%
\subsection{The Effect of B-Spline and Uncertainty Propagation}
\label{sec:Ablation}

%We conducted additional experiments on the module-wise contribution to the entire framework.
\textbf{(\textit{i}) Module-wise comparison: } Entire algorithm (\texttt{FULL}) consists of B-spline interpolation (\texttt{CNT}) and uncertainty propagation with localization weight (\texttt{UNC}). We test the baseline that exploits IMU discrete model and the same weight for all points (denoted as \texttt{RAW}). Additionally, we run a method that utilizes only one state covariance for uncertainty propagation, as in the case of M-LOAM (\texttt{F-UNC}). In this case, we exploit the last covariance of the scan as state covariance for point-wise uncertainty propagation. All of the methods are verified using ATE$_t$.

%TABLE
\begin{table}[t]
    \scriptsize
    \caption{Component-wise comparison using ATE$_t$}
    \label{table:ablation}
    \centering
    \resizebox{0.8\columnwidth}{!}{\begin{tabular}{cccccc}
        \toprule

                             & \multicolumn{1}{c}{\texttt{RAW}} & \multicolumn{1}{c}{\texttt{CNT}}                                       & \multicolumn{1}{c}{\texttt{F-UNC}}                            & \multicolumn{1}{c}{\texttt{UNC}}                                       & \multicolumn{1}{c}{\texttt{FULL}}                                      \\ \midrule[1pt]
\multicolumn{1}{c|}{\texttt{City01}}  & 7.345                         & 7.280         & 7.001 & \textit{6.831}  & \textbf{6.538} \\
\multicolumn{1}{c|}{\texttt{City02}}  & 7.346                         & \textit{6.835}       & 7.058 & 6.844  & \textbf{6.707}  \\
\multicolumn{1}{c|}{\texttt{City03}}  & 6.043                         & 5.969           & 6.547 & \textit{5.837}  & \textbf{5.470} \\ \midrule[1pt]
\multicolumn{1}{c|}{\texttt{Mongok}}  & 2.652                         & \textit{2.597}  & 2.645 & 2.611           & \textbf{2.579}  \\
\multicolumn{1}{c|}{\texttt{Whampoa}} & 4.728                         & 4.463           & 4.657 & \textbf{4.078} & \textit{4.236} \\
\multicolumn{1}{c|}{\texttt{TST}}     & 2.721                         & \textbf{2.143} & 2.773 & 2.752         & \textit{2.324} \\ \bottomrule
    \end{tabular}}
\vspace{-3mm}
    \end{table}

%TABLE

%FIGURE: ablation
\begin{figure}[!t]
  \centering
  \subfigure[The output of \texttt{RAW}]{%
	  \includegraphics[width=.47\columnwidth]{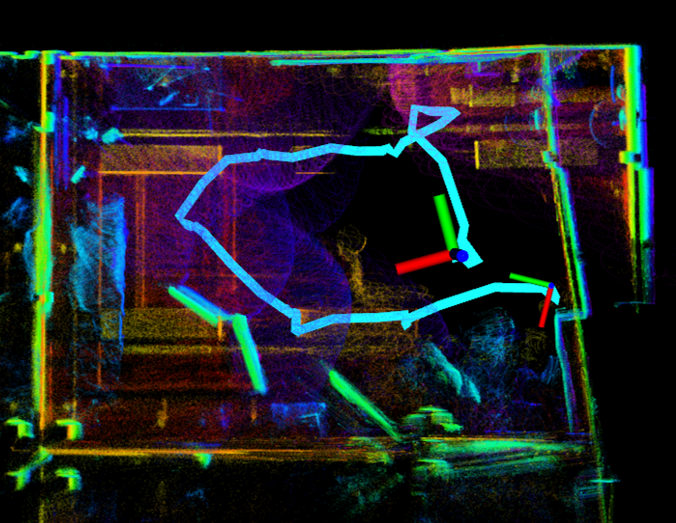}
    \label{fig:HiltiFast}
  }%
  \subfigure[The output of \texttt{UNC}]{%
	  \includegraphics[width=.47\columnwidth]{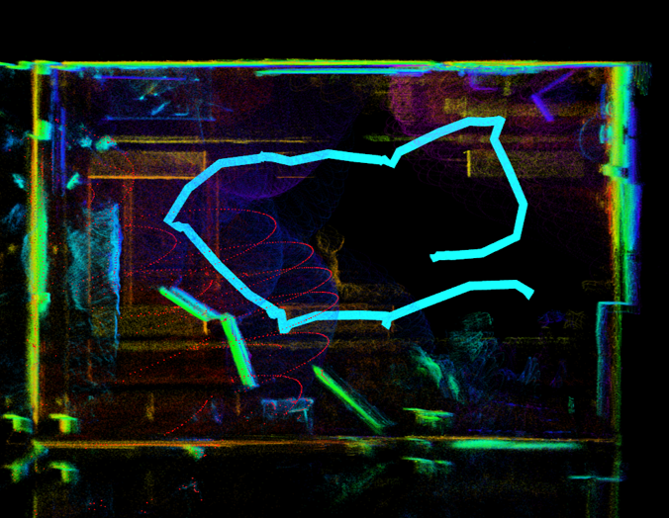}
    \label{fig:HiltiOurs}
  }
	\vspace{-2mm}
	\caption{This map is generated using Livox Horizon after the successful initialization (\unit{60}{sec} from starting) in the \texttt{LAB}. (a) unsuccessful mapping in this degenerate environment. (b) consistent map by the proposed method overcoming the challenges environment.}
  \label{fig:HiltiUncertainty}
  \vspace{-5mm}
\end{figure}
%FIGURE: ablation

\tabref{table:ablation} presents the ATE$_t$ for each test case. Both \texttt{RAW} and \texttt{CNT} address temporal discrepancies utilizing interpolation techniques; however, the \texttt{CNT} demonstrates slight improvement by using B-spline interpolation. This interpolation is particularly effective for datasets that exhibit highly curved trajectory. As expected, the ATE$_t$ is reduced in the UrbanNav dataset, which has a higher IMU frequency (\unit{400}{Hz}) than the our own Dataset (\unit{100}{Hz}). This result demonstrates that temporal compensation and undistortion with B-spline interpolation are effective in both cases, regardless of the IMU frequency. The effect of point-wise uncertainty is the most critical, as can be seen from \texttt{UNC}, while \texttt{F-UNC} shows some level of error reduction but not as much as ours. 

{In \texttt{City01-03}, \texttt{FULL} yields the best results by effectively harnessing the benefits of both methods. However, the UrbanNav dataset shows varied results due to environmental factors. Stationary and low-speed segments lead to less pronounced effects of B-spline interpolation and point-wise uncertainty, resulting in similar performance in \texttt{Mongok}. In \texttt{TST}, the inclined LiDAR occasionally detects fewer points and becomes the primary LiDAR, leading to lower uncertainty compared to the front LiDAR. Consequently, the performance of \texttt{UNC} slightly declines compared to \texttt{RAW}, which subsequently impacts the results of \texttt{FULL}. Conversely, in \texttt{Whampoa}, increased environmental complexity allows for more effective point-wise uncertainty and localization weight, even when the primary LiDAR changed, resulting in significant performance increase in \texttt{UNC}. Although the competing effect between B-spline interpolation and point-wise uncertainty leads to lower performance in \texttt{FULL} compared to \texttt{UNC}, the error is significantly reduced compared to \texttt{RAW}. In conclusion, both proposed methods generally improves performance when uses individually or in combination.}

\textbf{(\textit{ii}) Mapping capability: } A {qualitative} comparison is given in \figref{fig:HiltiUncertainty} when we test \texttt{RAW} and \texttt{UNC} in \texttt{LAB} sequence. Because the small \ac{FOV} is critically limited in this small indoor environment, this sequence is tricky to localize when equipped only with Livox Horizon. Unlike \texttt{RAW} that fails in consistent mapping, \texttt{UNC} localizes successfully in this confined space. As evidenced by the results in \tabref{table:ablation} and \figref{fig:HiltiUncertainty}, we can infer that the uncertainty model plays a leading factor in the enhancement of performance in our method.

\begin{figure}[!t]
	\centering
		\includegraphics[width=1\columnwidth]{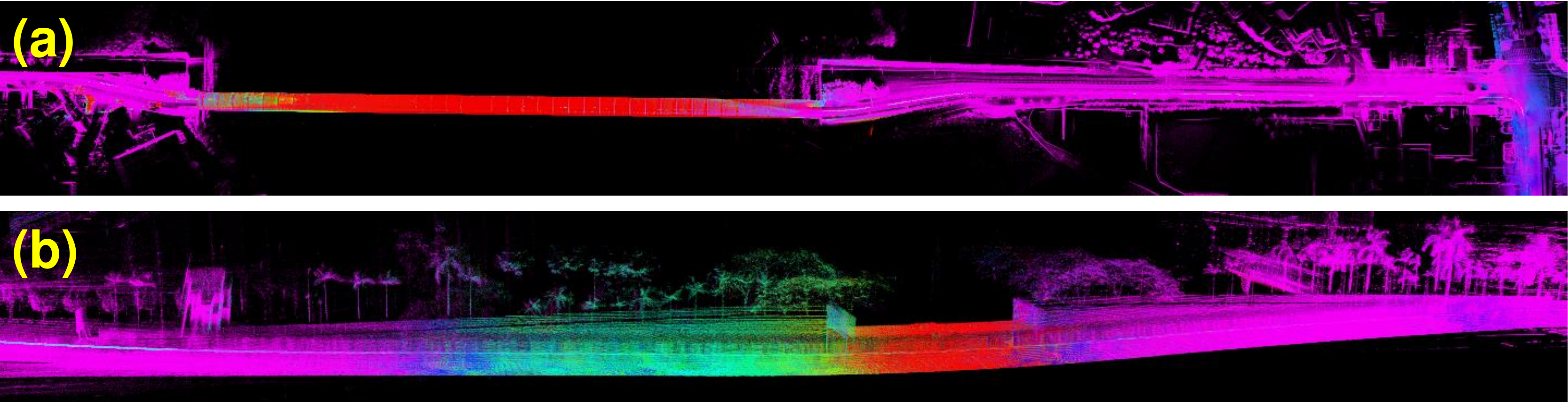}
  	\caption{Partial maps of (a) \texttt{City02} and (b) \texttt{Whampoa} with points color-coded by their localization weight from low (red) to high (purple). For both maps, a noticeable decrease in localization weight (red) only in the tunnel region. (b) the weight decrease is slighter than (a) because the upper part of the underpass is open and being scanned by inclined LiDAR.}
	\label{fig:tunnel}
	\vspace{-5mm}
\end{figure}

\textbf{(\textit{iii}) Localization weights: } In \figref{fig:tunnel}, we depict partial maps of the \texttt{City02} and \texttt{Whampoa}. The narrow nature of these environments can cause a correspondence error, leading to the influence of prior residuals being stronger than measurement residuals. As can be seen from the \figref{fig:tunnel}(a) and (b), localization weight is significantly reduced only at the degenerate surroundings. This reduction is expected and demonstrates the effectiveness of our localization weight in these challenging environments.

%FIGURE: localization weight
%FIGURE

%----------------------------------------------------------------------%
\subsection{The Effect of the Number of LiDAR}

\begin{figure}[!t]
  \centering
  \includegraphics[width=0.85\columnwidth]{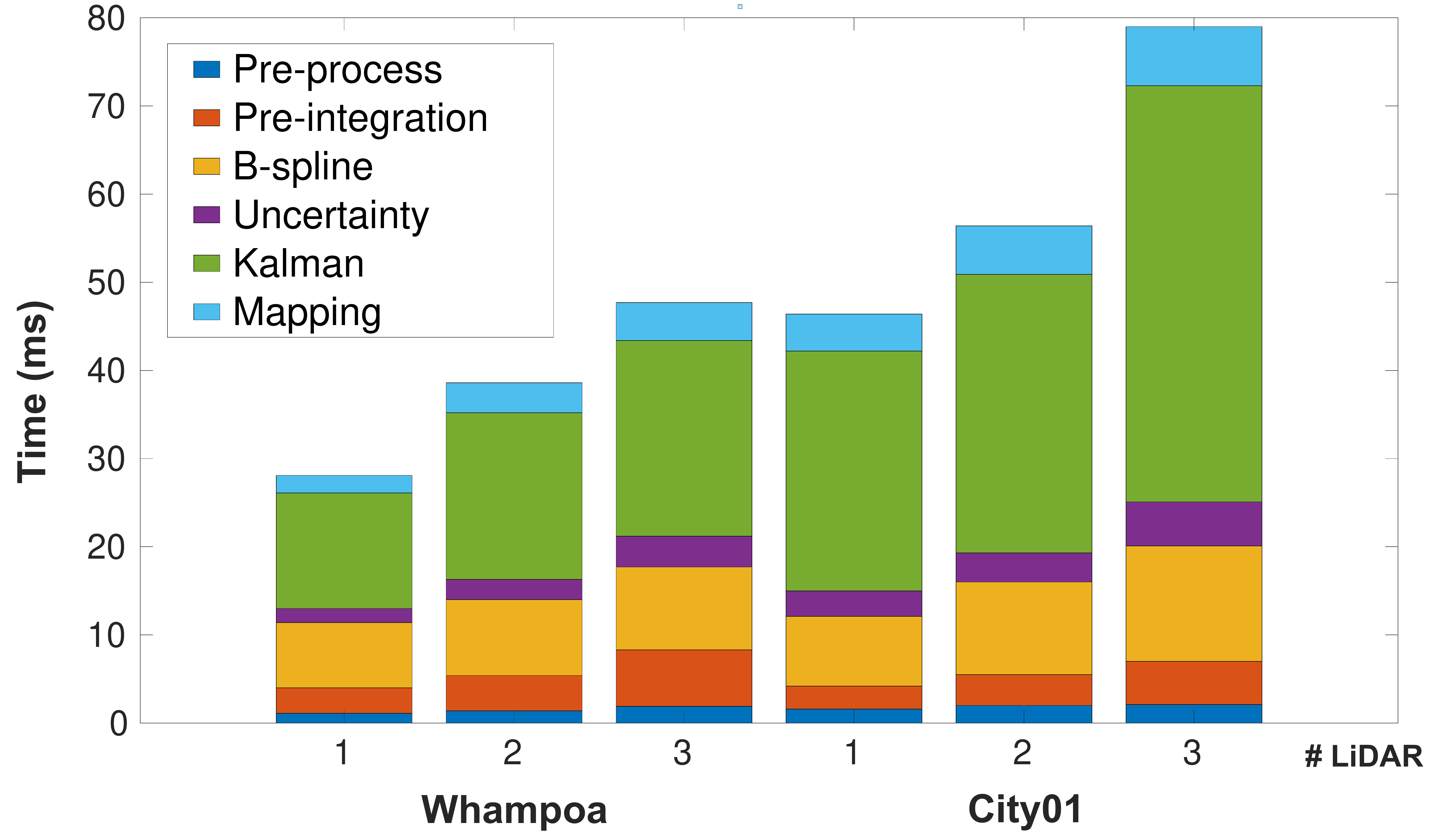}
  \vspace{-2mm}
  \caption{Computation times with respect to the number of LiDARs. \texttt{Pre-process} is the computation time for reformatting the pointcloud. \texttt{Pre-integration} involves integrating IMU measurements and undistorting the point. \texttt{B-spline} is utilized during undistortion and is separately listed from pre-integration. \texttt{Uncertainty} includes time for calculating the point-wise uncertainty and localization weight.}
  \label{fig:Time}
  \vspace{-6mm}
\end{figure}

%We check the effect of the number of LiDARs on time consumption and accuracy.
To assess the real-time performance, we measure and analyze the average processing time per scan for each module, as in \figref{fig:Time} and \tabref{table:timeConsumption}.

%FIGURE
%FIGURE

%TABLE
% Please add the following required packages to your document preamble:
% \usepackage{multirow}
\begin{table}[!b]
    \scriptsize
\vspace{-5mm}
\caption{Time analysis according to the number of LiDAR [ms]. The numbering follows the description in \tabref{tab:dataset}.}
    \label{table:timeConsumption}
    \centering
    \resizebox{0.48\textwidth}{!}{\begin{tabular}{ccccccc} \toprule
\multicolumn{1}{l}{}                  & \multicolumn{3}{c|}{\texttt{Whampoa}}            & \multicolumn{3}{c}{\texttt{City01}} \\ \midrule
\multicolumn{1}{c|}{LiDAR \#}         & 1    & 2    & 3                         & 1      & 2       & 3       \\
\multicolumn{1}{c|}{Point (Down.) \#} & 4045 & 5647 & \multicolumn{1}{c|}{6784} & 8554   & 10172   & 12244   \\
\multicolumn{1}{c|}{LiDAR} & HDL-32E & +VLP-16 & \multicolumn{1}{c|}{+LS-C16} & OS2-128   & +Avia   & +Tele   \\ \midrule[1pt]
\multicolumn{1}{c|}{Preprocess}       & 1.1  & 1.4  & \multicolumn{1}{c|}{1.9}  & 1.6    & 2.0     & 2.1     \\
\multicolumn{1}{c|}{Pre-integration}  & 2.9  & 4    & \multicolumn{1}{c|}{6.4}  & 2.6    & 3.5     & 4.9     \\
\multicolumn{1}{c|}{B-Spline}         & 7.4  & 8.6  & \multicolumn{1}{c|}{9.4}  & 7.9    & 10.5    & 13.1    \\
\multicolumn{1}{c|}{Uncertainty}      & 1.6  & 2.3  & \multicolumn{1}{c|}{3.5}  & 2.9    & 3.3     & 5       \\
\multicolumn{1}{c|}{Kalman Filter}    & 13.1 & 18.9 & \multicolumn{1}{c|}{22.2} & 27.2   & 31.6    & 47.2    \\
\multicolumn{1}{c|}{Mapping}          & 2.0  & 3.4  & \multicolumn{1}{c|}{4.3}  & 4.2    & 5.5     & 6.7     \\ \midrule[1pt]
\multicolumn{1}{c|}{Total}            & 28.1 & 38.6 & \multicolumn{1}{c|}{47.7} & 46.4   & 56.4    & 79    \\ \bottomrule
\end{tabular}}

\footnotesize
\raggedright
\vspace{0.8mm}
{*Specification: Intel i7 CPU@2.50Ghz and 48GB RAM} \\
\vspace{-7mm}
\end{table}

%TABLE

\textbf{(\textit{i}) Time: } We analyze the time consumption for \texttt{Whampoa} and \texttt{City01}, which have the most extensive distance among the UrbanNav and our datasets. {During all experiments, downsampling with $0.4$m resolution was implemented. \tabref{table:timeConsumption} exhibits the number of points obtained following this operation.} As can be seen, our method is light-weight and fully supports real-time performance even for a multi-LiDAR system. One of the modules in our method, uncertainty propagation, can be completed within a maximum of \unit{5}{ms}, even though it significantly improves performance. In contrast, the B-spline interpolation takes longer because undistortion is done without downsampling. However, even in \texttt{City01} with two Livox and an Ouster LiDAR, our method spends \unit{79}{ms} fully supporting \unit{10}{Hz}). Further improvement is possible by performing downsampling before the undistortion if needed.

\begin{figure}[!t]
  \centering
  \subfigure[ATE$_t$ in the \texttt{TST} and \texttt{Whampoa}]{%
    \includegraphics[width=0.55\columnwidth]{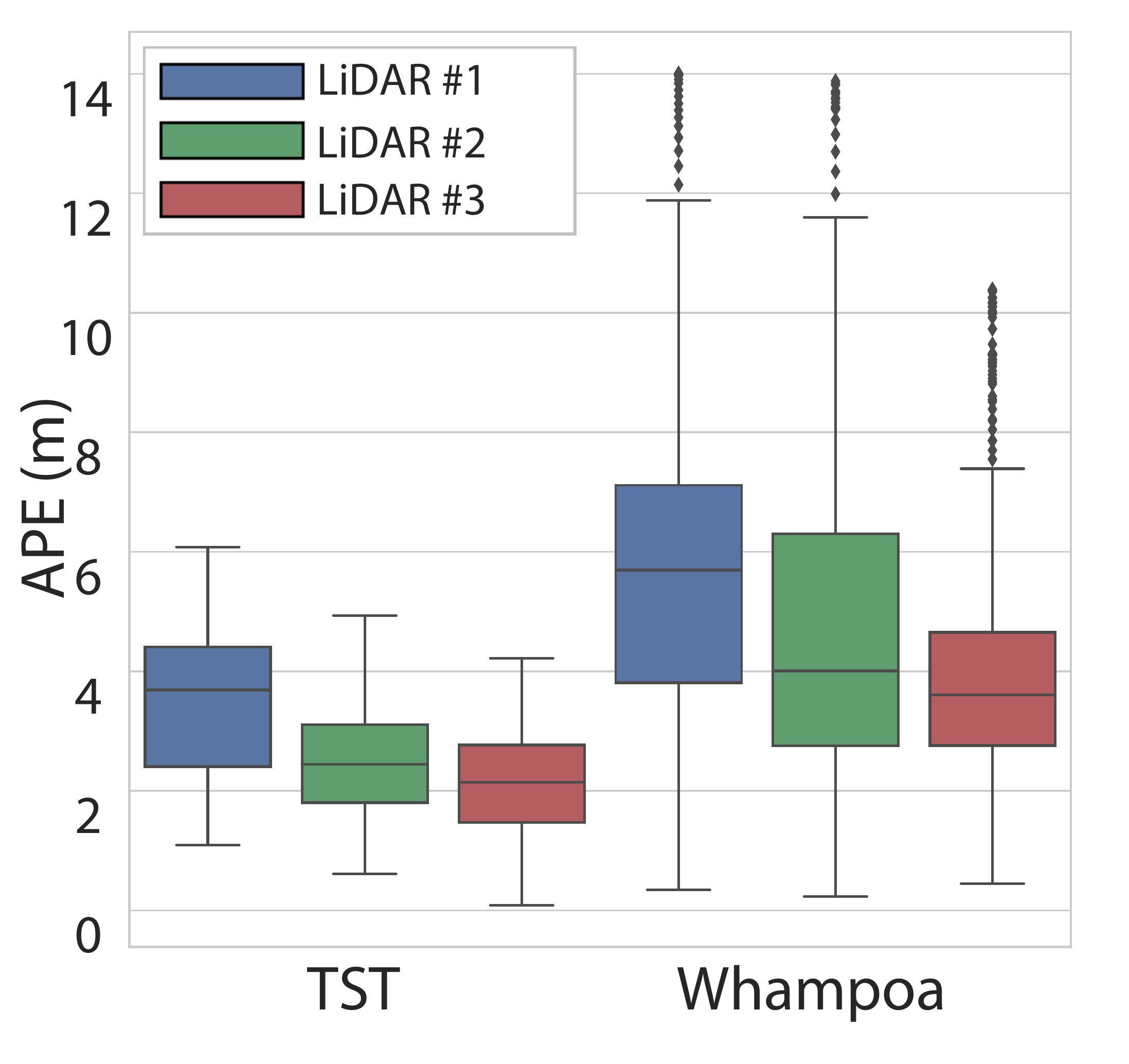}
    \label{fig:3lidar_cmp}
  }%
  \subfigure[\texttt{City03}]{%
    \begin{minipage}[b]{0.4\columnwidth} \centering
    \resizebox{\columnwidth}{!}{
    \begin{tabular}{l|c} \toprule
      No.               & Median ATE$_t$\\ \midrule
      1 (\texttt{RAW})  & 5.76   \\
      1 (\texttt{FULL}) & 5.65   \\
      2 (\texttt{RAW})  & 5.27   \\
      2 (\texttt{FULL}) & 5.08   \\
      3 (\texttt{RAW})  & 5.22   \\
      3 (\texttt{FULL}) & 4.73 \\
      \bottomrule
      \end{tabular}%
      }
      \vspace{10pt}%
    \end{minipage}
    \label{fig:3lidar_cmp_table}
  }
  \caption{\subref{fig:3lidar_cmp} an evaluation about three cases for two datasets. From left, the median values of ATE$_t$ and standard deviations are: (3.69, 1.22), (\textit{2.44}, \textbf{0.96}), (\textbf{2.14}, \textit{0.98}), (5.69, \textit{2.53}), (\textit{4.01}, 2.69) and (\textbf{3.58}, \textbf{1.93}). \subref{fig:3lidar_cmp_table} shows the median ATE$_t$ in \texttt{City03} as a function of the number of LiDARs for the \texttt{FULL} and \texttt{RAW}.}
  \label{fig:accuracy_number}
\vspace{-6mm}
\end{figure}

\textbf{(\textit{ii}) Accuracy: } The impact of the number of LiDAR on accuracy is further evaluated using three datasets: \texttt{TST}, \texttt{Whampoa}, and \texttt{City03}. For this test, we compare the results with \texttt{RAW} in \texttt{City03}. The errors decrease as the number of LiDAR increases, and the error range tends to decrease when checking the interquartile range and standard deviation in \figref{fig:accuracy_number}.

Interestingly, a significant error reduction is observed when the number increases from one to two, while a minor reduction is observed when the number increases from two to three (\figref{fig:accuracy_number}(a)). We speculate this is because an increment to two LiDARs provides sufficient constraints. Furthermore, as can be seen in the table in \figref{fig:accuracy_number}(b), it could be a fallacy to posit more LiDARs necessarily improve the performance. However, in \texttt{City03}, adding the second (Livox) and the third (Ouster) LiDARs could not effectively reduce the error because of the narrow \ac{FOV} and wide overlap of the two LiDAR. Using two LiDARs with \texttt{FULL} outperforms using three LiDARs with \texttt{RAW}. This finding highlights the need for careful consideration of LiDAR placement in a multi-LiDAR system. Simply merging the pointcloud in the overlapping area may not necessarily lead to significant improvements in performance. Leveraging B-spline interpolation to locate points accurately and assign the uncertainty, \texttt{FULL} with three LiDARs gives the best performance.

%Kalman filter is the time for solving the \ac{IESKF}. Finally, the mapping involves adding and managing points to the Ikd-tree.

\section{conclusion}
\label{sec:conclusion}

In this paper, we proposed a framework for asynchronous multiple LiDAR-inertial systems. The proposed framework utilizes B-spline interpolation in conjunction with an IMU discrete model to mitigate the temporal discrepancy among multiple LiDARs. To mitigate the accumulation of ambiguity during frame changes, a common issue in temporal compensation methods, we proposed a method for propagating point-wise uncertainty based on IMU acquisition time and point distance from sensors. Additionally, we incorporated a localization weight to improve performance in challenging environments. Our method was validated through extensive experimentation on both public datasets and our dataset and achieved real-time performance while surpassing the state-of-the-art in terms of accuracy and robustness.
%Lastly, we publicly open the algorithm to promote further research and development for the LiDAR community.

%FIGURE
%\begin{figure}[!t]
%  \centering
%  \subfigure[ATE$_t$ in the \texttt{TST} and \texttt{Whampoa}]{%
%	  \includegraphics[width=.47\columnwidth]{figure/Channel.pdf}
%    \label{fig:Urban_number}
%  }%
%  \subfigure[ATE$_t$ in the \texttt{City03}]{\includegraphics[width=.47\columnwidth]{figure/Channel.pdf}
%    \label{fig:Private_number}
%  }%
%  \caption{(a) We evaluated three cases for two datasets. From left, the median values of ATE$_t$ and standard deviations are: (3.69, 1.22), (\textit{2.44}, \textbf{0.96}), (\textbf{2.14}, \textit{0.98}), (5.69, \textit{2.53}), (\textit{4.01}, 2.69) and (\textbf{3.58}, \textbf{1.93}). (b) We evaluated four cases for \texttt{City03}, including \textbf{Raw} for three LiDAR. The median and standard deviations are (5.65, 3.36) (\textit{5.08}, 3.17), (5.22, \textit{2.63}) and (\textbf{4.73}, \textbf{2.60}).}
%  \label{fig:accuracy_number}
%\vspace{-5mm}
%\end{figure}
%FIGURE

\footnotesize
%\balance
\bibliographystyle{IEEEtranN}
\bibliography{string-short,root}

\end{document}